\documentclass[abstract,a4paper]{scrartcl}
\usepackage[algo2e,english,boxruled,noline,noend,linesnumbered]{algorithm2e}
\usepackage[english]{babel}
\usepackage{amssymb,latexsym,amsmath,stmaryrd}
\usepackage{amsthm}
\usepackage{multirow}
\usepackage{rotating}
\usepackage{threeparttable}
\usepackage{longtable}
\usepackage[usenames,dvipsnames]{color}
\usepackage{fancyvrb}
\usepackage{epsfig}

\usepackage[all]{xy}

\usepackage{comment}

\includecomment{WithAppendix}

\newcommand{\ignore}[1]{}
\newcommand{\pjs}[1]%
{\textcolor{blue}{[[#1]]}\marginpar{\textcolor{blue}{\textsc{PJS}}}}
\newcommand{\mw}[1]%
{\marginpar{\textcolor{OliveGreen}%
{\textsc{Mark}}}\textcolor{OliveGreen}{[[#1]]}}
\newcommand{\as}[1]%
{\marginpar{\textcolor{red}{\textsc{Andreas}}}\textcolor{red}{[[#1]]}}

\newcommand{\Delete}[1]{\textcolor{red}{#1}}

\renewcommand{\Delete}[1]{}

\SetKw{KwAll}{all}
\SetKw{KwAnd}{and}
\SetKw{KwDownTo}{downto}
\SetKw{KwOr}{or}
\SetKw{KwTo}{to}

\newcommand{\cO}{\mathcal{O}} 
\newcommand{\CC}{\mathcal{C}} 
\newcommand{\DD}{\mathcal{D}} 

\newcommand{\EE}{{\cal E}}
\newcommand{\RR}{{\cal R}}
\newcommand{\VV}{{\cal V}}

\newcommand{\lit}[1]{\llbracket #1 \rrbracket}

\newcommand{\UB}{U\!B}

\newcommand{\vsids}{\textsc{Vsids}}
\newcommand{\restart}{\textsc{Restart}}

\newcommand{\rcpsp}{\textsc{Rcpsp}}
\newcommand{\rcpspmax}{\textsc{Rcpsp}/max}

\newcommand{\mslf}{\textsc{Mslf}}
\newcommand{\HS}{\textsc{Hot Start}}
\newcommand{\HR}{\textsc{Hot Restart}}

\newcommand{\cumu}{\texttt{cumulative}}

\newcommand{\FontBenchmark}[1]{\textsc{#1}}
\newcommand{\BCc}{\FontBenchmark{c}}
\newcommand{\BCd}{\FontBenchmark{d}}
\newcommand{\BCj}[1]{\FontBenchmark{j#1}}
\newcommand{\BCubo}[1]{\FontBenchmark{ubo#1}}

\newcommand{\subscript}[1]{\ensuremath{_{\textrm{#1}}}}
\newcommand{\bbs}{B\&B\subscript{S98}}
\newcommand{\bbdorn}{B\&B\subscript{D00}}
\newcommand{\evaball}{\textsc{Eva}}
\newcommand{\ises}{\textsc{Ises}}
\newcommand{\ifs}{\textsc{Ifs}}
\newcommand{\ifsfr}{\textsc{Ifs-Fr}}
\newcommand{\ifsmcsr}{\textsc{Ifs-Mcsr}}
\newcommand{\swo}{\textsc{Swo(br)}}
\newcommand{\fbsfel}{\textsc{Fbs}\subscript{F01}}
\newcommand{\dmfel}{\textsc{Dm}\subscript{F01}}
\newcommand{\gafel}{\textsc{Ga}\subscript{F01}}

\newcommand{\Dom}[2]{[#1..#2]}

\DeclareMathOperator{\domain}{\mathit{domain}}

\DeclareMathOperator{\DOM}{\mathit{DOM}}
\DeclareMathOperator{\false}{\mathit{false}}
\DeclareMathOperator{\true}{\mathit{true}}

\DeclareMathOperator{\vars}{\mathit{vars}}

\DeclareMathOperator{\solv}{\mathit{solv}}

\newcommand{\range}[2]{\left[\,#1\,..\,#2\,\right]}

\newcommand{\before}{\ll}

\newcommand{\institute}[1]{\publishers{\small #1}}
\newcommand{\email}[1]{\textsf{#1}}

\newtheorem*{acknowledgements}{Acknowledgements}

\theoremstyle{definition}
\newtheorem{example}{Example}

\begin{document}

\title{Solving the Resource Constrained Project Scheduling Problem with
Generalized Precedences by Lazy Clause Generation}

\author{
   Andreas Schutt$^\dagger$ \and
   Thibaut Feydy$^\dagger$ \and
   Peter J. Stuckey$^\dagger$ \and
   Mark G. Wallace$^\star$
}

\institute{
   $^\dagger$~National ICT Australia, Department of Computer Science \& Software
Engineering, The~University~of~Melbourne, Victoria 3010, Australia\\
   \email{\{aschutt,tfeydy,pjs\}@csse.unimelb.edu.au}\\[1em]
   $^\star$~Faculty of Information Technology,\\ Monash University, Clayton, Vic 3800, Australia\\
   \email{mark.wallace@infotech.monash.edu.au}
}

\date{}
\maketitle

\begin{abstract}
   The technical report presents a generic exact solution approach for
   minimizing the project duration of the resource-constrained project
   scheduling problem with generalized precedences (\rcpspmax{}).
   The approach uses lazy clause generation, i.e., a hybrid of finite domain
   and Boolean satisfiability solving, in order to apply nogood
   learning and conflict-driven search on the solution generation.
   Our experiments show the benefit of lazy clause generation for finding an
   optimal solutions and proving its optimality in comparison to other 
   state-of-the-art exact and non-exact methods.
   The method is highly robust: it matched or bettered
   the best known results on all of the 2340 
   instances we examined except 3, according to the 
   currently available data on the PSPLib.
   Of the 631 open instances in this set it closed 573 and improved the
   bounds of 51 of the remaining 58 instances.
\end{abstract}


\section{Introduction}


The Resource-constrained Project Scheduling Problem with generalized precedences
(\rcpspmax)\footnote{In the literature \rcpspmax{} is also called as \rcpsp{}
with temporal precedences, arbitrary precedences, minimal and maximal time lags,
and time windows.} consists of scarce resources, activities and precedence
constraints between pairs of activities.
Each activity requires some units of resources
during their execution.
The aim is to build a schedule that obeys the resource and precedence
constraints.
Here, we concentrate on renewable resources (i.e., their supply is constant
during the planning period), non-preemptive activities (i.e. once started their
execution cannot be interrupted), and finding a schedule that minimizes the
project duration (also called \textit{makespan}).
This problem is denoted as $PS\vert temp\vert C_{max}$ by
Brucker et al.~\cite{Brucker:etal:99}
and $m, 1\vert gpr\vert C_{max}$ by Herroelen et al.~\cite{Herroelen:etal:98}.
Bartusch et al.~\cite{Bartusch:etal:88} show that the decision wether an instance is feasible or not is already NP-hard.
\ignore{Bartusch et al.~\cite{Bartusch:etal:88} show that the decision variant
of this problem is already NP-hard.}

The \rcpspmax{} problem is widely studied and some of its applications can be
found in Bartusch et al.~\cite{Bartusch:etal:88}.
A problem instance 
consists of 
a set of resources, 
a set of activities, and a set of
generalized precedences between activities.
Each resource is characterized by its
discrete capacity, and each 
activity by its discrete processing time (duration)
and its resource requirements.
Generalized precedences express relations of start-to-start, start-to-end, end-
to-start, and end-to-end times between pairs of activities.
All these relations can be formulated as start-to-start times precedence.
Those precedences have the form $S_i + d_{ij} \leq S_j$ where $S_i$ and $S_j$
are the start times of the activities $i$ and $j$ resp., and $d_{ij}$ is
a discrete distance between them.
If $d_{ij}$ is non-negative this imposes a minimal time lag, 
while if $d_{ij}$ is negative this imposes a maximal time lag between start
times.

\begin{figure}
\centerline{
\includegraphics[width=15cm]{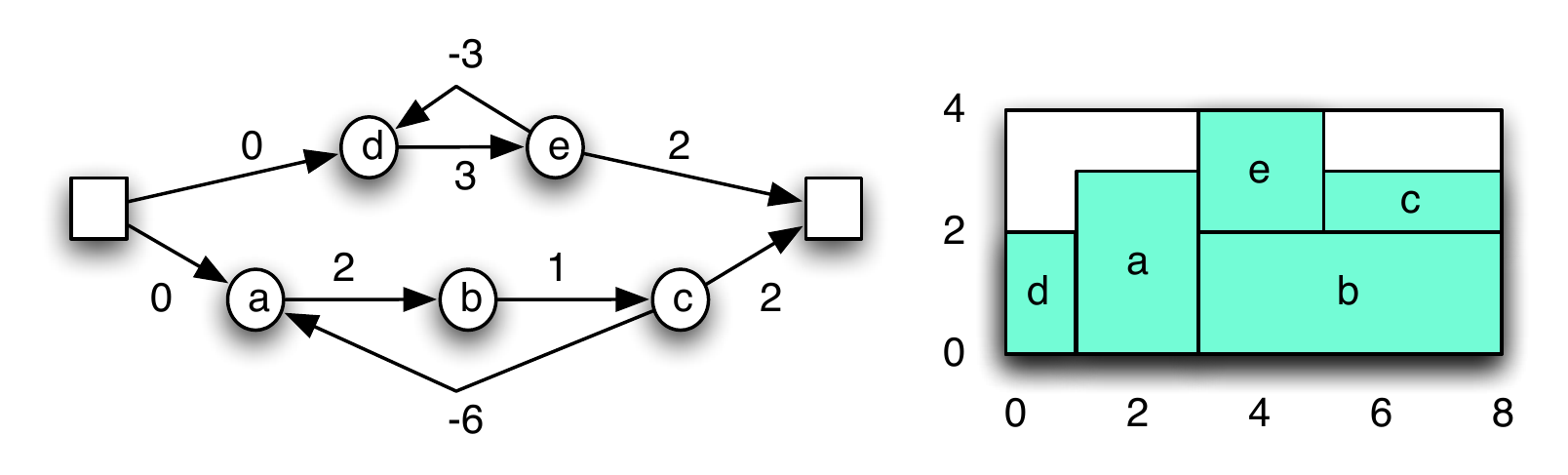}}
\caption{The precedence graph and a solution to a small \rcpspmax{} problem.
\label{fig:rcpsp}}
\end{figure}

\begin{example}
   \label{ex:cumu}
   A simple example of an \rcpspmax{} problem consists of the five activities $[a, b, c, d, e]$ with their start times $[s_a, s_b, s_c, s_d, s_e]$, their durations $[2, 5, 3, 1, 2]$ and resource requirements on a single resource $[3, 2, 1, 2, 2]$ and a resource capacity of 4.
   Suppose we also have the generalized precedences $s_a + 2 \leq s_b$ 
   (activity~$a$ ends before activity~$b$ starts), $s_b + 1 \leq s_c$ 
   (activity~$b$ starts at least 1 time unit before activity~$c$ starts), $s_c - 6 \leq s_a$ (activity~$c$ can not start later than 6 time units after activity~$a$ starts), $s_d + 3 \leq s_e$ (activity~$d$ starts at least 3 time units before activity~$e$ starts), and $s_e - 3 \leq s_d$ (activity~$e$ can not start later than 3 time units after activity~$d$ starts).
   Note that the last two precedences express the relation $s_d + 3 = s_e$ 
   (activity~$d$ starts exactly 3 time units before activity~$e$).

   Let the planning horizon, in which all activities must be completed be
   8. Figure~\ref{fig:rcpsp} illustrates the precedence graph between the
five tasks and source at the left (time 0) and sink at the right (time 8),
as well as a potential solution to this problem, 
where a rectangle for activity~$i$ has width equal to its duration and height equal to its resource requirements.
\hfill $\Box$
\end{example}

To our knowledge the first exact method to tackle \rcpspmax{} was proposed by
Bartusch et al.~\cite{Bartusch:etal:88}. 
They use a branch-and-bound algorithm
to tackle the problem. Their branching is based on resolving (minimal) conflict
sets\footnote{Conflict sets are set of activities for which their execution
might overlap in time and violate at least one resource constraint if they are
executed at the same time.} by the addition of precedence constraints breaking
these sets.
Later other branch-and-bound methods were developed which are based on the same
idea (e.g. De Reyck and Herreolen~\cite{DeReyck:Herroelen:98}, 
Schwindt~\cite{Schwindt:98}, and Fest et al.~\cite{Fest:etal:99}). The results from Schwindt are the best published one for an exact method on the testset \textsc{Sm} so far.

Dorndorf et al.~\cite{Dorndorf:etal:00} use a time-oriented branch-and-bound
combined with constraint propagation for precedence and resource constraints. In
every branch one unscheduled and ``eligible'' activity is selected and its start
time is assigned to the earliest point in time that does not violate any
constraint regarding the current partial schedule.
In the case of backtracking they apply dominance rules to fathom search space.
As far as we can determine this 
exact approach outperforms other exact methods for
\rcpspmax{} on the \textsc{CD} benchmark set.

Franck et al.~\cite{Franck:etal:01} compare different solution methods on a
benchmark set \textsc{UBO} with instances ranging from 10 to 1000 activities.
Their methods are a truncated branch-and-bound algorithms, filter-beam search,
heuristics with priority rules, genetic algorithms and tabu search.
All methods share a preprocessing step to 
determining feasibility or infeasibility.
The preprocessing step 
decomposes the precedence network into strongly connected components (SCCs)
(which are denoted ``cyclic structures'' in~\cite{Franck:etal:01}). 
The preprocessing then 
determines a solution or infeasibility for each SCC individually
using constraint propagation and a destructive lower bound
computation.
Once a solution for all SCCs is determined a first solution can be
deterministically generated for 
the original instance; otherwise infeasibility is proven.
\ignore{
Their results are the best published results on their benchmark set 
\textsc{UBO} for
instances up to 100 activities.}

Ballest{\'i}n et al.~\cite{Ballestin:etal:09} employ an evolutionary algorithm based on a serial generation scheme with unscheduling step. Their crossover operator is based on so called \emph{conglomerates}, i.e. set of cycle structures and other activities which cannot move freely inside a schedule, it tries to keep the ``good'' conglomerates of the parents to their children.
This is the best published local search method so far on the testsets \textsc{UBO} (up to instances with 100 activties) and \textsc{CD}.

Cesta at al.~\cite{Cesta:etal:02} propose a two layered heuristic that is
based on a temporal precedence network and extension of this network by new
temporal precedence in order to resolve minimal conflict sets.
For guidance, constraint propagation algorithms are applied on the network.
Their method is competitive on the benchmark set \textsc{SM}.

Oddi and Rasconi~\cite{Oddi:Rasconi:09} apply a generic iterative search consisting of  of a relaxation and flatting step based on temporal precedences which are used for resolving resource conflicts. In the first step some of the temporal precedences are removed from the problem and then in the second others added if a resource conflict exists. 
Their methods is evaluate on instances with 200 activities from the benchmark set \textsc{UBO}.

A special case of \rcpspmax{} is the well-studied Resource-constrained Project
Scheduling Problem (\rcpsp{}) where the precedence constraints
$S_i + d_{ij} \leq S_j$ express that the activity $j$ must start after the end
of $i$, i.e. $d_{ij}$ equals to the duration of $i$.
In contrast to \rcpspmax{} the decision variant of \rcpsp{} is polynomial
solvable, but not its optimization variant which is NP-hard
(Blazewicz et al.~\cite{Blazewicz:etal:83}).
The reason for this is the absence of maximal time lags, i.e. here activity
executions can always be delayed 
until to a point in time where enough resource units
are available without breaking any precedence constraints. That is not possible
for \rcpspmax{}.

The best exact methods for \rcpsp{} to our knowledge 
are our own~\cite{Schutt:etal:09,Schutt:etal:10} and
Horbach~\cite{Horbach:10}. Both
use advanced SAT technology in order to take advantage of its nogood learning
facilities.
Our methods~\cite{Schutt:etal:09,Schutt:etal:10} are a generic approach based on
the Lazy Clause Generation (LCG)~\cite{Ohrimenko:etal:09} using 
the G12 Constraint
Programming Platform~\cite{G12:05}. Lazy clause generation 
is a hybrid of a finite domain and a
Boolean satisfiability solver.
Our approaches model 
the cumulative resource constraint either by decomposing into smaller
primitive constraints, or by
the creating a global \cumu{} propagator.
The global propagation approach performs better as the size of
the problem grows.
In contrast to our methods Horbach's approach is a hand-tailored for \rcpsp{},
but similarly a hybrid with SAT solving. He uses a linear programming
solver to determine activity schedules and hybridize with the SAT solver.
Overall our global approach~\cite{Schutt:etal:10} is superior
to Horbach's approach on \rcpsp{}.

In this paper  
we apply the same generic lazy clause generation approach 
to the more general problem of \rcpspmax{}.
Because the problems are more difficult than pure RCPSP
we need to modify our approach in particular to prove
feasibility/infeasibility
We show that the approach to solving \rcpspmax{} 
performs better than published methods so far, especially for
improving a solution, once a solution is found, and proving optimality.
We state the limitations of our current model and how to overcome them.
We compare out approach to the best known approaches to
\rcpspmax{} on several  benchmark suites 
accessible via the PSPLib~\cite{psplib}.

The paper is organized as follows. In Section~\ref{sec:prelim}
we give an introduction to lazy clause generation.
In Section~\ref{sec:model} we present our basic model for \rcpspmax{}
and discuss some improvements to it.
In Section~\ref{sec:search} we discuss the various branch-and-bound
procedures
that we use to search for optimal solutions.
In Section~\ref{sec:exp} we compare our algorithm to the best approaches
we are aware of on 3 challenging benchmark suites.
Finally in Section~\ref{sec:conc} we conclude.

\section{Preliminaries}
\label{sec:prelim}

In this section we explain lazy clause generation by first introducing
finite domain propagation and DPLL based SAT solving, and then explaining
the hybrid approach. We discuss how the hybrid explains conflicts and
briefly discuss how a \texttt{cumulative} propagator is extended to
explain its propagations.

\subsection{Finite Domain Propagation}

Finite domain propagation (see e.g.~\cite{toplas09}) is a powerful approach 
to tackling combinatorial problems.
A finite domain problem ($\CC$,$D$) 
consists of a set of constraints $\CC$ over a set of variables
$\VV$, a domain $D$ which determine the finite set of possible values of
each variable in $\VV$.
A \emph{domain} $\DD$ is a complete mapping from $\VV$ to finite sets of
integers.
Hence given domain $D$, then $D(x)$ is the set of possible values that
variable $x$ can take. Let $\min_D(x) = \min (D(x))$ 
and $\max_D(x) = \max(D(x))$.
Let $\range{l}{u} = \{ d \mid l \leq d \leq u, d \in \mathbb{Z} \}$ 
denote a \emph{range} of integers., where $\range{l}{u} = \emptyset$ if $l > u$.
In this paper we will concentrate on domains where 
$D(x)$ is a range for all $x \in \VV$.
The \emph{initial domain} is referred as $D_{init}$.
Let $D_1$ and $D_2$ be domains, then $D_1$
is \emph{stronger} than $D_2$, written $D_1 \sqsubseteq D_2$, if
$D_1(v) \subseteq D_2(v)$ for all $v \in \VV$.
Similarly if $D_1 \sqsubseteq D_2$ then $D_2$ is \emph{weaker} than $D_1$.
For instance, all domains $D$ that occur will be stronger than the initial
domain, i.e.~$D \sqsubseteq D_{init}$.

A \emph{valuation} $\theta$ is a mapping of variables
to values, written $\{x_1 \mapsto d_1, \ldots, x_n \mapsto d_n\}$.
We extend the valuation $\theta$ to map expressions or constraints
involving the variables in the natural way.  Let $\vars$ be the
function that returns the set of variables appearing in an expression,
constraint or valuation.
In an abuse of notation, we define a valuation $\theta$ to be an
element of a domain $D$, written $\theta \in D$, if $\theta(v) \in
D(v)$ for all $v \in \vars(\theta)$.
Define a \emph{valuation domain} $D$ as one where $|D(v)| = 1, \forall v \in
\VV$. We can define the corresponding valuation $\theta_D$ for a valuation
domain $D$ as $\{ v \mapsto d ~|~ D(v) = \{d\}, v \in \VV \}$.

Then a constraint $c\in \CC$ is a set of valuations over $vars(c)$ which give
the allowable values for a set of variables.
In FD solvers constraints are implemented by propagators.
A \emph{propagator}~$f$ implementing ~$c$ is a monotonically decreasing
function on domains such that for all domains $D \sqsubseteq
D_{init}$:
$f(D) \sqsubseteq D$ and
no solutions are lost, i.e.
$\{\theta\in D \mid \theta \in c\} = \{\theta \in f(D)\mid \theta \in c\}$.
We assume each propagator $f$ is \emph{checking}, that is if $D$ is a valuation
domain then $f(D) = D$ iff $\theta_D$ restricted to $vars(c)$
is a solution of $c$. 
Given a set of constraints $\CC$ we assume a corresponding set of propagators
$F = \{ f ~|~ c \in \CC, f \text{~implements~} c\}$.

A \emph{propagation solver} for a set of propagators $F$ and
current domain $D$, $\solv(F,D)$, repeatedly applies all the propagators
in $F$ starting from domain $D$ until there is no
further change in resulting domain.
$\solv(F,D)$ is the weakest domain $D' \sqsubseteq D$ which is
a fixpoint (i.e.~$f(D') = D'$) for all $f \in F$.
\ignore{
\mw{Surely we are not claiming our propagators reach the fixpoint??}
\pjs{no, how do you interpret it this way!}
}

Finite domain solving interleaves propagation with search decisions.
Given a initial problem ($\CC$,$D$)
where $F$ are the propagators for the constraints $\CC$ we
first run the propagation solver $D' = \solv(F, D)$. 
If this determines failure then the problem has no solution
and we backtrack to visit the next unexplored choice.
If $D$ is a valuation domain we have determined a solution.
Otherwise we pick a variable $x \in \VV$ and split its domain $D'(x)$
into two disjoint parts $S_1 \cup S_2 = D'(x)$ 
creating two subproblems 
$(\CC, D_1)$, $(\CC, D_2)$,
where $D_i(x) = S_i$ and $D_i(v) = D'(v), v \neq x$, 
whose solutions are the solutions of the original problem.
We then recursively explore the first problem, 
and when we have shown it has no solutions
we explore the second problem.

As defined above finite domain propagation is only applicable to
\emph{satisfaction problems}.  Finite domain solvers solve optimization
problems by mapping them to repeated satisfaction problems. Given an
objective function $o$ to minimize under constraints $\CC$ with domain $D$,
the finite domain solving approach first finds a solution
$\theta$ to $(\CC, D)$, and then finds a solution  to
$(\CC \cup \{ o \leq \theta(o)\}, D)$, that is, the satisfaction problem of
finding a better solution than previously founds. It repeats this process
until a problem is reached with no solution, in which case the last found
solution is optimal.  If the process is halted before proving optimality,
then the solving process just returns the last solution found as the best
known.

Finite domain propagation is a powerful generic approach to solving
combinatorial optimization problems.  Its chief strengths are the ability to
model problems at a very high level, and the use of global propagators,
specialized propagation algorithms for important constraints.

\subsubsection{Cumulative}

Of particular interest to us in this work is the 
global \texttt{cumulative}
constraint for cumulative resource scheduling. 

The \cumu{} constraint introduced by Aggoun and
Beldiceanu~\cite{Aggoun:Beldiceanu:93} in 1993
is a constraint with Zinc~\cite{Zinc:08} type
\begin{quote}\tt\small
\begin{tabbing}
predicate cumulative(\=list of var int: s, list of \=int: d,\\
\> list of int: r, \>int: c);
\end{tabbing}
\end{quote}
Each of the first three arguments are lists of the same length $n$
and indicate information about a set of \emph{activities}.
$s[i]$ is the \emph{variable} \emph{start time} of the $i^{th}$ activity,
$d[i]$ is the fixed \emph{duration} of the $i^{th}$ activity,
and
$r[i]$ is the fixed \emph{resource usage} (per time unit) of the $i^{th}$ activity.
The last argument $c$ is the fixed \emph{resource capacity}.

The \cumu{} constraints represent \cumu{} resources with a constant
capacity over the considered planning horizon applied to non-preemptive activities,
\emph{i.e.} if they are started they cannot be interrupted.
Without loss of generality we assume that all values are integral and non-negative and there is a
\emph{planning horizon} $t_{max}$ which is the latest time any activity can
finish.

\begin{example}
   \label{ex:cumu:cons}
   Consider the five activities $[a, b, c, d, e]$ from Example~\ref{ex:cumu} with durations $[2, 5, 3, 1, 2]$ and resource requirements $[3, 2, 1, 2, 2]$ and a resource capacity of 4.
   This is represented by the \cumu{} constraint.
   $$
   \texttt{cumulative}([s_a,s_b,s_c,s_d,s_e],[2,5,3,1,2],[3,2,1,2,2].4)
   $$
   Imagine each task must start at time 0 or after and finish before time 8.
   The cumulative problem corresponds to packing the rectangles shown in
   Figure~\ref{fig:cumul}(a) into the resource box shown below.
   \hfill $\Box$
\end{example}
\begin{figure}
\begin{tabular}{cc}
\multicolumn{2}{c}{\includegraphics[width=13cm]{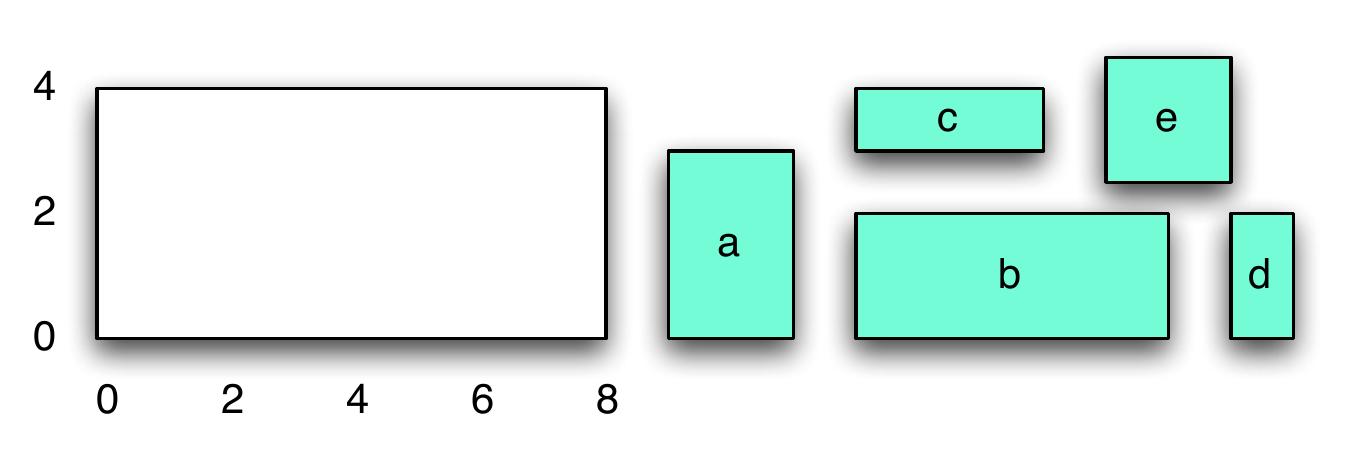}} \\
\multicolumn{2}{c}{(a)} \\
\includegraphics[width=6cm]{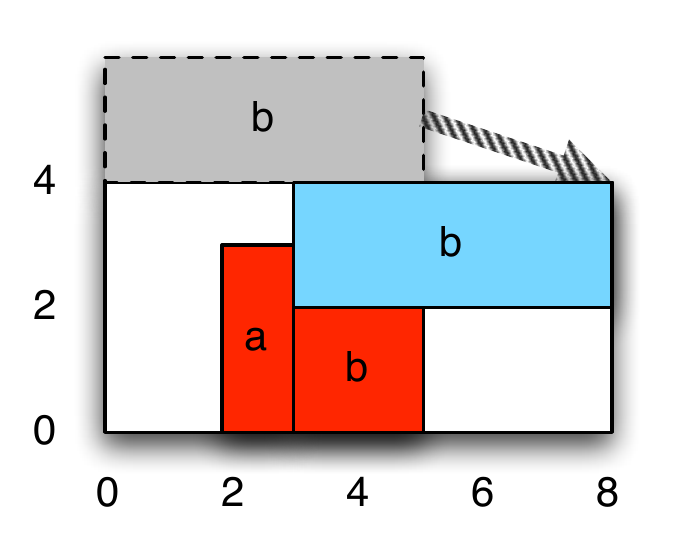} &
\includegraphics[width=6cm]{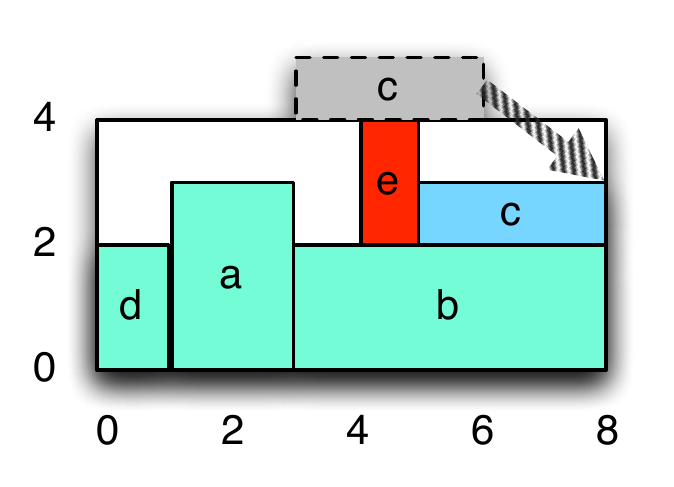} 
\\
(b) & (c) 
\end{tabular}
\caption{Figure illustrating propagation of 
the \cumu{} constraint.\label{fig:cumul}}
\end{figure}

There are many propagation algorithms for \cumu{}, but the most widely used
is based on timetable propagation~\cite{LePape:94}. 
An activity $i$ has a \emph{compulsory part} given domain $D$ from 
$[\max_D s[i] .. \min_D s[i] + d[i] -1]$, 
that requires that activity $i$
makes us of $r[i]$ resources at each of the times in
$\range{\max_D s[i]}{\min_D s[i] + d[i] - 1}$
if the range is non-empty. The timetable propagator
for \cumu{} first determines the \emph{resource usage} {profile} $ru[t]$ 
which sums for each time $t$ the resources requires for
all compulsory parts of activities at that time.
If at some time $t$ the profile 
exceeds the resource capacity, i.e. $ru[t] > c$, 
the constraint is violated and failure detected.
If at some time $t$ point the resource used in the profile 
that there is not enough left for an activity $i$, i.e. $ru[t] + r[i] > c$,
then we can determine that activity $i$ cannot be scheduled to
run during time $t$. If the earliest start time of activity $i$, $\min_D s[i]$
means that the activity cannot be scheduled completely before time $t$,
i.e. $\min_D s[i] + d[i] > t$, we
can update the earliest start time to be $t + 1$, similarly if the 
lastest start time of the activity means that the activity cannot be
scheduled completely after $t$ , i.e. $\max_D s[i] \leq t$ then we can
update the latest start time to be $t - d[i]$.
For a full description of timetable propagation 
for \cumu{} see e.g.~\cite{Schutt:etal:10}.

\begin{example}
   \label{ex:cumul:prop}
   Consider the \cumu{} constraint of Example~\ref{ex:cumu:cons}
   where the domains of the start times are now
   $D(s_a) = \range{1}{2}$,
   $D(s_b) = \range{0}{3}$,
   $D(s_c) = \range{3}{5}$,
   $D(s_d) = \range{0}{2}$,
   $D(s_e) = \range{0}{4}$.
   Then there is compulsory parts of activities~$a$ and~$b$ in the ranges
   $[2..2]$ and $[3..4]$ respectively shown in
   Figure~\ref{fig:cumul}(b) in red. No other activities have a compulsory
   part. Hence the red contour illustrates the resource usage profile.
   Since activity~$b$ cannot be scheduled in parallel with activity~$a$,
   and the earliest start time of activity~$b$, 0, means that the activity
   cannot be scheduled before activity~$a$ we can reduce the domain of the
   start time for activity~$b$ to the singleton $\range{3}{3}$.
   This is illustrated in Figure~\ref{fig:cumul}(b).
   The opposite holds for activity~$a$ that cannot be run after activity~$b$, hence the domain of its start time shrinks to the singleton range $\range{1}{1}$.
   Once we make these changes the compulsory parts of the activities~$a$
   and~$b$ increase to the ranges $\range{1}{2}$ and $\range{3}{7}$
   respectively. This in turn causes the start times of activities~$d$
   and~$e$ to become $\range{0}{0}$ and $\range{3}{4}$ respectively creating
   compulsory parts in the ranges $\range{0}{0}$ and $\range{4}{4}$
   respectively.
   The later causes the start time of activity~$c$ to become fixed at $5$
   generating the compulsory part in $\range{5}{7}$ which causes that the
   start time of activity~$e$ becomes fixed at $3$.
   This is illustrated in Figure~\ref{fig:cumul}(c).
   In this case the timetable propagation results in a 
   final schedule in the right of Figure~\ref{fig:rcpsp}.
   \hfill $\Box$
\end{example}

\subsection{Boolean Satisfiability}

Let $\cal B$ be a set of Boolean variables. 
A \emph{literal} $l$ is Boolean variable $b \in {\cal B}$ or its negation
$\neg b$.
The negation of a literal $\neg l$ is defined as
$\neg b$ if $l = b$ and $b$ if $l \equiv \neg b$. 
A \emph{clause} $C$ is a set of Boolean variables understood as a disjunction.
Hence clause $\{l_1, \ldots, l_n\}$ is satisfied if at least one literal
$l_i$
is true. 
An \emph{assignment} $A$ is a set of Boolean literals that does not
include a variable and its negation, i.e. $\nexists b \in {\cal B}. \{b,
\neg b\} \subseteq A$. An assignment can be seen as a partial valuation
on Boolean variables, $\{ b \mapsto \true | b \in A\} 
\cup \{ b \mapsto \false | \neg b \in A\}$.
A theory $T$ is a set of clauses.
A \emph{SAT problem} $(T,A)$ consists of a set of clauses $T$
and an assignment over (some of) the variables occuring in $T$.

A Davis-Putnam-Loveland-Logemann (DPLL) SAT solver is a form of finite
domain propation solver specialized for Boolean clauses. 
Each clause is propagated by so called \emph{unit propagation}.
Given an assignment $A$, unit propagation detects failure
using clause $C$ is such that
$\{ \neg l ~|~ l \in C\} \subseteq A$, and unit propagation detects
a new unit consequence $l$ if $C \equiv \{l\} \cup C'$ and
$\{ \neg l' ~|~ l' \in C'\} \subseteq A$, in which case it adds
$l$ to the current assignment $A$.
Unit propagation continues until failure is detected, or no new unit
consequences can be determined. 

SAT solvers exhaustively 
apply unit propagation to the current assignment $A$ to generate
all the consequences possible $A'$. 
They then choose an unfixed Boolean variable $b$ and create two equivalent
problem $(T, A' \cup \{b\})$, $(T, A' \cup \{\neg b\})$ and recursively
search these subproblems.  The Boolean literals added to the
assignment by choice are termed \emph{decision literals}.

Modern DPLL based SAT solving is a powerful approach to solving combinatorial
optimization problems because it records nogoods that prevent the search
from revisiting similar parts of the search space. 
The SAT solver records an explanation for each unit consequence discovered
(the clause that caused unit propagation), and on failure uses these
explanations to determine a set of mutually incompatible decisions, a
\emph{nogood}  which is added as a new clause to the theory of the problem. 
These nogoods drastically reduce
the size of the search space needed to be examined.  Another advantage of
SAT solvers is that they track which Boolean variables are involved in the
most failures (called \emph{active} variables), 
and use a powerful autonomous search procedure which
concentrates on the variables that are most active.  The disadvantages of SAT
solvers are the restriction to Boolean variables and the sometime huge
models that are required to represent a problem because the only constraints
expressible are clauses.

\subsection{Lazy Clause Generation}

Lazy clause generation is a hybrid of finite domain propagation and
Boolean satisfiability.   The key idea in lazy clause generation is
to run a finite domain propagation solver, but to build explanation of
the propagations made by the solver by recording them as clauses
on a Boolean variable representation of the problem. Hence as the FD search
progresses we lazily create a clausal representation of the problem.
The hybrid has the advantages of FD solving, but inherits the SAT solvers
ability to create nogoods to drastically reduce search, and use activity
based search.

\subsubsection{Variable Representation}
\label{ssec:varrepr}

\ignore{
\mw{In this paper we employ a somewhat novel syntax to denote boolean variables.  Instead of naming each variable with a letter, we name it according to its meaning or interpretation in the problem.  To indicate that it is a boolean variable we use the syntax $\lit{.}$.  For example $\lit{x=d}$ is the Boolean variable whose truth is interpreted as meaning that the finite domain variable $x$ has the value $d$. Naturally its falsity means that $x\neq d$.
(The correctness of this interpretation is normally enforced through
explicit constraints.)}
}

In lazy clause generation 
each integer variable $x\in\VV$ with the initial domain
$D_{init}=\range{l}{u}$ is represented by the following Boolean variables
$\lit{x = l}, \ldots, \lit{x = u}$ and
$\lit{x \leq l}, \ldots, \lit{x \leq u - 1}$.
The variable $\lit{x = d}$ is true if $x$ takes the value $d$, and false if $x$
takes a value different from $d$.
Similarly the variable $\lit{x \leq d}$ is true if $x$ takes a value less than
or equal to $d$ and false for a value greater than $d$.
Note that we use $\lit{x= d}$ and $\lit{x \leq d}$ throughout the paper as
the \emph{names} of Boolean variables.
Sometimes the notation $\lit{d \leq x}$ is used for the literal
$\neg \lit{x \leq d - 1}$.

Not every assignment of Boolean variables is consistent with the integer
variable~$x$, for example $\{ \lit{x = 3}, \lit{x \leq 2} \}$ (i.e. both Boolean
variables are true) requires that $x$ is both 3 and $\leq 2$.
In order to ensure that assignments represent a consistent set of possibilities
for the integer variable $x$ we add to the SAT solver the clauses $\DOM(x)$ that
encode
$\lit{x \leq d} \rightarrow \lit{x \leq d+1}, l \leq d < u$,
$\lit{x = l} \leftrightarrow \lit{x \leq l}$,
$\lit{x = d} \leftrightarrow (\lit{x \leq d} \wedge \neg \lit{x \leq
d- 1}), l < d < u$, and
$\lit{x = u} \leftrightarrow \neg \lit{x \leq u-1}$
where $D_{init}(x) = \range{l}{u}$.
We let $\DOM = \cup \{ \DOM(v) ~|~ v \in \VV \}$.

Any assignment $A$ on these Boolean variables can be converted to a domain:
$
domain(A)(x) = \{  d \in D_{init}(x) \mid \forall \lit{c} \in A,
  vars(\lit{c}) = \{x\}: x = d \models c\},
$
that is, the domain includes all values for $x$ that are consistent with all the
Boolean variables related to $x$.
It should be noted that the domain may assign no values to some variable.

\begin{example}
   \label{ex:assign}
   Consider Example~\ref{ex:cumu} and assume $D_{init}(s_i) =
   \range{0}{15}$ for $i\in \{a, b, c, d, e\}$.  The assignment$A = \{ \neg
   \lit{s_a \leq 1}, \neg \lit{s_a = 3}, \neg \lit{s_a = 4}, \lit{s_a \leq
     6}, \neg \lit{s_b \leq 2}, \lit{s_b \leq 5}, \neg \lit{s_c \leq 4},
   \lit{s_c \leq 7}, \neg \lit{s_e \leq 3} \}$ is
   consistent with $s_a = 2$, $s_a = 5$, and $s_a = 6$. Hence
   $domain(A)(s_a) = \{2,5,6\}$.  For the remaining variables
   $domain(A)(s_b) = \range{3}{5}$, 
   $domain(A)(s_c) = \range{5}{7}$,
   $domain(A)(s_d) = \range{0}{15}$, and $domain(A)(s_e) = \range{4}{15}$.
   Note that for brevity $A$ is not a fixpoint of a SAT propagator for
   $DOM(s_a)$ since we are missing many implied literals such as $\neg
   \lit{s_a = 0}$, $\neg \lit{s_a = 8}$, $\neg \lit{s_a \leq 0}$ etc.
   \hfill $\Box$
\end{example}

\subsubsection{Explaining Propagators}

In LCG a propagator is extended from a mapping from domains to domains to a
generator of clauses describing propagation.
When $f(D) \neq D$ we assume the propagator $f$ can determine a clause
$C$ to explain each domain change.
Similarly when $f(D)$ is a false domain the propagator must create a clause
$C$ that explains the failure.

\begin{example}
   \label{ex:diffcons}
   Consider the propagator $f$ for the precedence constraint $s_a + 2 \leq s_b$ from Example~\ref{ex:cumu}.
   When applied to the domains $D(s_i) = \range{0}{15}$ for $i\in \{a,b\}$ it obtains $f(D)(s_a) = \range{0}{13}$, and $f(D)(s_b) = \range{2}{13}$.
   The clausal explanation of the change in domain of $s_a$ is
   $\lit{s_b \leq 15} \rightarrow \lit{s_a \leq 13}$,
   similarly the change in domain of $s_b$ is $\neg \lit{s_a \leq -1} \to \neg \lit{s_b \leq 1}$ 
   ($\lit{s_a \geq 0} \to \lit{s_b \geq 2}$).
   These become the clauses $\neg \lit{s_b \leq 15} \vee \lit{s_a \leq 13}$
   and $\lit{s_a \leq -1} \vee \neg \lit{s_b \leq 1}$.
   \hfill $\Box$
\end{example}

The explaining clauses of the propagation are added 
to the database of the SAT
solver on which unit propagation is performed.
Because the clauses will always have the form $C \rightarrow l$ where
$C$ is a conjunction of literals true in the current assignment,
and $l$ is a literal not true in the current assignmet,
the newly added clause will always cause unit propagation,
adding $l$ to the current assignment.

\begin{example}
   \label{ex:diffcons2}
   Consider the propagation from Example~\ref{ex:diffcons}. The clauses
   $\neg \lit{s_b \leq 15} \vee \lit{s_a \leq 13}$ and
   $\lit{s_a \leq -1} \vee \neg \lit{s_b \leq 1}$ are added to the SAT theory.
   Unit propagation infers that $\lit{s_a \leq 13} = \true$ and
   $\neg \lit{s_b \leq 1} = \true$ since $\neg \lit{s_b \leq 15}$ and
   $\lit{s_a \leq -1}$ are $\false$, and adds these literals to the assignment.
   Note that the unit propagation is not finished, since for example 
   the implied literal $\lit{s_a \leq 14}$, can be detected $\true$ as well. 
\hfill $\Box$
\end{example}

The unit propagation on the added clauses $C$ is 
guaranteed to be as strong as the propagator~$f$ on
the original domains, i.e. if $\domain(A)\sqsubseteq D$ then
$\domain(A') \sqsubseteq f(D)$ where $A'$ is the resulting assignment after
addition of $C$ and unit propagation (see~\cite{Ohrimenko:etal:09} for the formal
results).

Note that a single new propagation may be explainable
using different set of clauses.
In order to get maximum benefit from the explanation we desire a ``strongest''
explanation as possible.
A set of clauses $C_1$ is \emph{stronger} than a set of clauses $C_2$ if $C_2$
implies $C_1$.
In other words, $C_1$ restricts the search space at least as much as $C_2$.

\begin{example}
   \label{ex:cumulex}
   Consider explaining the propagation of the start time of the activity~$c$
   described in Example~\ref{ex:cumul:prop} and Figure~\ref{fig:cumul}(c).
   The domain change $\lit{5 \leq s_c}$
   arises from the compulsory parts of activity $b$ 
   and $e$    as well as the fact that activity~$c$ 
   cannot start before time 3.
   An explanation of the propagation is hence 
$\lit{3\leq s_c} \wedge \lit{3\leq s_b} \wedge \lit{s_b\leq 3} \wedge \lit{3\leq s_e} \wedge \lit{s_e \leq 4} \to \lit{5 \leq s_c}$.
   
   We can observe that if $2\leq s_c$ 
then the same domain change $\lit{5\leq s_c}$ follows due to the compulsory parts of activity~$b$ and~$e$.
   Therefore, a stronger explanation is obtained by replacing the literal $\lit{3\leq s_c}$ by $\lit{2\leq s_c}$.
   
   Moreover, the compulsory parts of the activity~$b$ in the ranges
   $\range{3}{3}$ and $\range{5}{7}$ are not necessary for the domain
   change. We only require that there is not enough resources at time 4
   to schedule task $c$.  
   Thus the refined explanation can be further strengthened by
   replacing 
   $\lit{3\leq s_b} \wedge \lit{s_b\leq 3}$ by
   $\lit{s_b\leq 4}$ which is enough to force a compusory part of $s_b$
   at time 4. This leads to the stronger 
   explanation $\lit{2\leq s_c}
   \wedge \lit{s_b\leq 4} \wedge \lit{3\leq s_e}
   \wedge \lit{s_e \leq 4} \to \lit{5 \leq s_c}$.
\end{example}

In this example the final explanation corresponds to a pointwise explanation
in Schutt et al.~\cite{Schutt:etal:10}. Here, those pointwise explanations
are used to explain the timetable propagation.  For a full discussion about
the best way to explain \cumu{} propagation see~\cite{Schutt:etal:10}.

\subsubsection{Nogood generation}

Since all the propagation steps in lazy clause generation have been mapped
to unit propagation on clauses, we can perform nogood generation
just as in a SAT solver.  

The nogood generation is based on an \emph{implication graph} and the 
\emph{first unique implication point} (1UIP).
The graph is a directed graph where nodes represent fixed literals and directed
edges reasons why a literal became $\true$, and is extended as the search
progresses. Each unit propagation marks the literal it makes true
with the clause that caused the unit propagation. The true literals are kept
in a stack showing the order that they were determined as true by unit
consequence or decisions.

For clarity purpose, we do not differentiate between literal and node.
A literal is fixed either by a search decision or unit propagation.
In the first case the graph is extended only by the literal and in the second
case by the literal and incoming edges to that literal from all other literals
in the clause on that the unit propagation assigned the $\true$ value to the
literal.

\begin{example}
  Consider the strongest explanation $\lit{2\leq s_c} \wedge \lit{s_b\leq 4}
  \wedge \lit{3\leq s_e} \wedge \lit{s_e \leq 4} \to \lit{5 \leq s_c}$ from
  Example~\ref{ex:cumulex}. It is added to the SAT database as clause $\neg
  \lit{2\leq s_c} \vee \neg \lit{s_b\leq 4} \vee \neg\lit{3\leq s_e} \vee
  \neg\lit{s_e \leq 4} \vee \lit{5 \leq s_c}$ and unit propagation sets
  $\lit{5 \leq s_c}$ $\true$.  Therefore the implication graph is extended by
  the edges $\lit{2\leq s_c} \to \lit{5 \leq s_c}$, $\lit{s_b\leq 4} \to \lit{5 \leq s_c}$, $\lit{3\leq s_e}
  \to \lit{5 \leq s_c}$, and $\lit{s_e\leq 4} \to \lit{5 \leq s_c}$.  \hfill
  $\Box$
\end{example}

Every node and edge is associated with the search level at which they are added
to the graph.
Once a conflict encounters a nogood which is the 1UIP in LCG is calculated based
on the implication graph.
A conflict is recognized when the unit propagation reaches a clause where all literals are false. 
This clause is the starting point of the analysis and builds a first tentative nogood.
Now, literals in the tentative nogood are replaced by the literals from its
incoming edges in the reverse order of the graph extension.
This process holds on until the tentative nogood includes 
exactly one literal from the current decision level.
The resulting nogood is called \emph{1UIP} (first unique implication point),
since it corresponds to a cut through the implication graph that 
has one node in the current decision level.

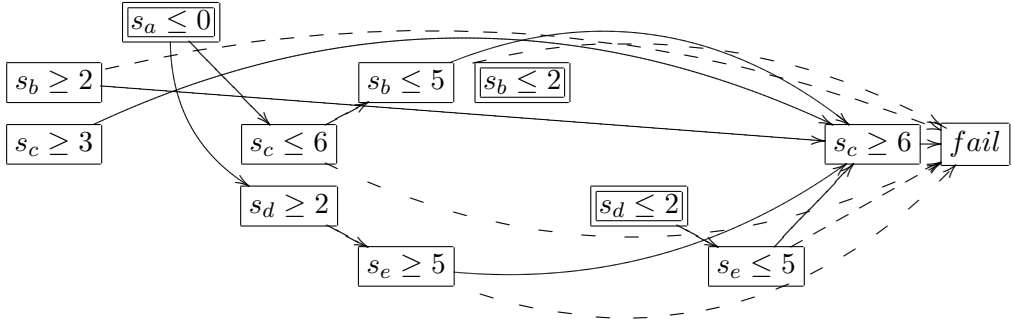
\begin{figure}
$$
    \newcommand{\xyo}[1]{*+[F-]{#1}}    
    \newcommand{\xyd}[1]{*+[F=]{#1}}    
    \newcommand{\xyor}[1]{*[red]+[F:red:-]{#1}}
\xymatrix@R=8pt@C=8pt{
&\xyd{s_a \leq 0} \ar[rdd] \ar@/_15pt/[rddd] \\
\xyo{s_b \geq 2} \ar[rrrrrrrd] \ar@{-->}@/^30pt/[rrrrrrrrd]&& & \xyo{s_b \leq 5} \ar@/^30pt/[rrrrd]\ar@{-->}@/^25pt/[rrrrrd] & \xyd{s_b \leq 2} \\
\xyo{s_c \geq 3} \ar@/^40pt/[rrrrrrr]&& \xyo{s_c \leq 6} \ar[ru]\ar@{-->}@/_35pt/[rrrrrr] &&&&& \xyo{s_c \geq 6} \ar[r] & \xyor{fail}    \\
&& \xyo{s_d \geq 2} \ar[rd] &&& \xyd{s_d \leq 2} \ar[rd] \\
&& & \xyo{s_e \geq 5} \ar@/_20pt/[rrrruu] \ar@{-->}@/_40pt/[rrrrruu] & & & \xyo{s_e \leq 5} \ar[ruu]\ar@{-->}[rruu] 
}
$$
\label{fig:conflict}
\caption{(Part of) The implication graph for the propagation of
  Example~\ref{ex:cumu:nogood}. Decision literals are shown double boxed, while
  literals set by unit propagation are shown boxed.}
\end{figure}

\begin{example}
   \label{ex:cumu:nogood}
   Considered the \rcpspmax{} instance from Example~\ref{ex:cumu} on 
   page~\pageref{ex:cumu}.
   
   Assume an initial domain of $D_{init} = \range{0}{15}$ then after the initial propagation of the precedence constraints the domains are $D(s_a) = \range{0}{8}$, $D(s_b) = \range{2}{10}$, $D(s_c) = \range{3}{12}$, $D(s_d) = \range{0}{10}$, and $D(s_b) = \range{3}{13}$.
   Note that no tighter bounds can be inferred by the cumulative propagator.

   Assume search now sets $s_a \leq 0$.
   This sets the literal $\lit{s_a \leq 0}$ as true, and  
   unit propagation on the domain clauses will set 
   $\lit{s_a = 0}$, $\lit{s_a \leq 1}$, $\lit{s_a \leq 2}$, etc.
   In the remainder of the example we will ignore propagation of the
   domain clauses and concentrate on the ``interesting propagation''.
   
   The precedence constraint $s_c -6 \leq s_a$ will force $s_c \leq 6$
   with explanation $\lit{s_a \leq 0} \to \lit{s_c \leq 6}$. 
   The the precedence constraint $s_b + 1 \leq s_c$ will force $s_b \leq 5$
   with explanation $\lit{s_c \leq 6} \to \lit{s_b \leq 5}$. 

   The timetable propagator for \cumu{} will use the compulsory part of
   activity $a$ in $[0..2)$ to force $s_d \geq 2$.  The explanation for this
   is $\lit{s_a \leq 0} \to \lit{s_d \geq 2}$. 
   The the precedence $s_d + 3 \leq s_e$ forces $s_e \geq 5$ with
   explanation  $\lit{s_d \geq 2} \to \lit{s_e \geq 5}$. 
   
   Suppose next that search sets $s_b \leq 2$. There is no propagation from
   precedence constraints or the \cumu{} constraint.
   It does create a compulsory part of $s_b$ from $[2..7)$ but there 
   is no propagation.

   Suppose now that search sets $s_d \leq 2$. Then the precedence constraint
   $s_e - 3 \leq s_d$ forces $s_e \leq 5$ with explanation
   $\lit{s_d \leq 2} \to \lit{s_e \leq 5}$. This creates a compulsory part
   of $d$ in $[2..3)$ and a compulsory part of $e$ in $[5..7)$.
   In fact all the activities $a$, $b$, $d$ and $e$ are fixed now.
   Timetable propagation sees since all resources are used at time 5 
   then activity $c$ cannot start before time 6. A 
   reason for this is $\lit{s_b \geq 2} \wedge \lit{s_b \leq 5}$ (which forces 
   $b$ to use 2 resources in [5..7)), plus 
   $\lit{s_e \geq 5} \wedge \lit{s_e \leq 5}$ (which forces 
   $e$ to use 2 resources in [5..7)), plus $\lit{s_c \geq 3}$ 
  (which forces $c$ to
   overlap this time. Hence an explanation is 
   $\lit{s_b \geq 2} \wedge \lit{s_b \leq 5}
    \wedge \lit{s_e \geq 5} \wedge \lit{s_e \leq 5}
    \wedge \lit{s_c \geq 3} \to \lit{s_c \geq 6}$.

   This forces a compulsory part of $c$ at time 6 which causes a resource
   overload at that time.  An explanation of the failure is
    $\lit{s_b \geq 2} \wedge \lit{s_b \leq 5}
    \wedge \lit{s_e \geq 5} \wedge \lit{s_e \leq 5}
    \wedge \lit{s_c \geq 6} \wedge \lit{s_c \leq 6} \to fail$. 
    The edges are shown in the conflict graph as dashed (for clarity).

    The nogood generation process starts from this original explanation.
   It removes the last literal in the explanation by replacing it by its
   explanation. Replacing  $\lit{s_c \geq 6}$ by its explanation creates the
   new nogood 
   $\lit{s_b \geq 2} \wedge \lit{s_b \leq 5}
    \wedge \lit{s_e \geq 5} \wedge \lit{s_e \leq 5}
    \wedge \lit{s_c \geq 3} \wedge \lit{s_c \leq 6} \to fail$. 
   Since this nogood has only one literal that was made true after the last
  decision level $\lit{s_e \leq 5}$ this is the 1UIP nogood. 
  Rewritten as a clause it is
  $\lit{s_b \leq 1} \vee \neg \lit{s_b \leq 5} \vee 
   \lit{s_e \leq 4} \vee \neg \lit{s_e \leq 5} \vee \lit{s_c \leq 2} \vee
   \neg \lit{s_c \leq 6}$.
\hfill $\Box$
\end{example}

After discovering a new nogood $C$ the 
lazy clause generation solver, like a SAT solver,
adds the clause $C$ to the theory, and backtracks to the decision level
of the second newest literal in  nogood $C$.  At this point we are
guaranteed
that the clause will unit propagate.  After unit propagation finishes search
proceeds as usual.

\begin{example}
Continuing Example~\ref{ex:cumu:nogood},
the  solver backtracks to the decision level of the second newest literal
(in  this case $s_e \leq 5$) thus undoing the decisions
$s_d \leq 2$ and $s_b \leq 2$ and their consequences.
The newly added nogood unit propagates to force $s_e \geq 6$
   with explanation $\lit{s_b \geq 2} \wedge
  \lit{s_b \leq 5} \wedge \lit{s_e \geq 5}
    \wedge \lit{s_c \geq 3} \wedge \lit{s_c \leq 6} \to \lit{s_e \geq 6}$,
   and the
   precedence constraint $s_e - 3 \leq s_d$ forces $s_d \geq 3$ with explanation
   $\lit{s_e \geq 6} \to \lit{s_d \geq 3}$. Search proceeds looking for a
   solution. 
\hfill $\Box$
\end{example}

\section{Models for RCPSP/max}
\label{sec:model}

In this section a basic model for \rcpspmax{} instance is presented at first,
and then different possible model improvements which are mainly based on
activities in disjunction.

An \rcpspmax{} problem can be represented as follows: A set of activities
$\VV=\{1, \dots, n\}$ is subjected to generalized precedences in
$\EE\subset \VV^2\times \mathbb{Z}$ between two activities, and scarce resources
in $\RR$.
The goal is to find a schedule $S = (S_i)_{i\in \VV}$ that respects the
precedence and resource constraints, and minimizes the project duration 
(makespan) where $S_i$ is the \emph{start time} of the activity~$i$.

Each activity~$i$ has a finite \emph{processing time} or \emph{duration}~$p_i$
and requires (non-negative) $r_{ik}$ units of resource $k$, $k\in \RR$ 
for its execution where $r_{ik}$ is the \emph{resource requirement}
or \emph{usage} of activity $i$ for resource $k$. 
A resource~$k\in \RR$ has a
constant\footnote{Variation of resource capacities can be obtained by using
artificial activities that claim the not-available resource units.}
\emph{resource capacity}~$R_k$ over the planning period which cannot be exceeded
at any point in time. The planning period is given by $[0, t_{max})$ where
$t_{max}$ is the maximal planning horizon.

Generalized precedences $(i,j, d_{ij})\in \EE$ between the activities $i$ and
$j$ are subjected to the constraint $S_i + d_{ij} \leq S_j$, i.e. it represents
a \emph{minimal time lag} ($j$ must start at least $d_{ij}$ time units after $i$
starts) if $d_{ij} \geq 0$ and a \emph{maximal time lag} ($i$ must start at most
$-d_{ij}$ time units after the start of $j$) if $d_{ij} < 0$.
Generalized precedences encode not only start-to-start relations between
activities, but also start-to-end, end-to-start, and end-to-end by addition/
subtraction of $i$'s or $j$'s duration to $d_{ij}$.
If a minimal time lag $d_{ij}^{+}$ and a maximal time lag $d_{ji}^{-}$ exist for
an activity $j$ concerning to $i$ then the start time $S_j$ is restricted to
$\Dom{S_i + d_{ij}^{+}}{S_i - d_{ji}^{-}}$.
In the case of $d_{ij}^{+} = -d_{ji}^{-}$ the activity $j$ must start exactly
$d_{ij}^{+}$ time units after $i$.

For the remainder of this section let
an \rcpspmax{} instance be given with activities $\VV = \{1, 2, \dots, n\}$,
generalized precedences~$\EE$, resources~$\RR$, and a planning period~
$[0, t_{max})$. Then the basic model can be stated as the following
Zinc~\cite{Zinc:08} model.
\begin{quote}\tt\small\centering
\begin{tabbing}
\%---------------------------------------------------------------------\%\\
\% Parameters\\
int: t\_max; ~~~~~ \=\% The planning horizon \\
set of int: R; \>\% The set of resources\\
set of int: V; \>\% The set of activities\\
set of int: Idx; \>\% The index set of precedences\\
\\
array [Idx, 1..3] of \=int: rcap; \=\% The precedences of form x + c <= y\kill
array [R] of \>int: rcap; \>\% The resource capacities\\
array [V] of \>int: p; \>\% The activities durations\\
array [V, R] of \>int: r; \>\% The activities resource requirements\\
array [Idx, 1..3] of \>int: E; \>\% The precedences of form x + c <= y\\
\\
set of int: Times = 0..t\_max; \% The planning period\\
mmmm\=mmmm\=mmmm\=mmmm\=\kill
\%---------------------------------------------------------------------\%\\
\% Variables\\
array [V] of var Times: S;\\
var Times: objective;\\
\%---------------------------------------------------------------------\%\\
\% Constraints\\
\> \% Precedence constraints\\
constraint\\
\> forall (id in Idx) (S[E[id, 1]] + E[id, 2] <= S[E[id, 3]]);\\
\\
\> \% Cumulative resource constraints\\
constraint\\
\> forall (res in R) (cumulative(S, p, [r[i, res] | i in V], rcap[res]));\\
\\
\> \% Objective constraints\\
constraint\\
\>   forall (i in V) (S[i] + p[i] <= objective);\\
\%---------------------------------------------------------------------\%\\
\% Search\\
solve minimize objective;\\
\%---------------------------------------------------------------------\%\\
\end{tabbing}
\end{quote}

This basic model has a number of weaknesses: 
first the initial domains of the start
times are large, second each precedence constraint is modelled as one individual
propagator, and finally the SAT solver in LCG has no structural information
about activities in disjunction.

A smaller initial domain can be computed by taking into
account the precedences in $\EE$
as described in the next subsection. Individual propagators for precedences may
not be so bad for a small number of precedences, but for a larger number of
propagators, their queuing behaviour may result in long 
and costly sequences of propagation steps.
A global propagator can efficiently adjust the time-bounds in $\cO(n\log n + m)$
time as described in Feydy et al.~\cite{Feydy:etal:08},
but we did not have access to such a propagator for the experiments.
Reified precedence constraints can be used for modelling activities in
disjunctions as described later in this section.

\subsection{Initial Domain}
\label{ssec:initdomain}

A smaller initial domain can be obtained 
for the start time variables 
by applying the Bellman-Ford single source
shortest path algorithm~\cite{Bellman:58,Ford:Fulkerson:62} on the digraph
$G=(\VV', \EE')$ where $\VV' = \VV\cup\{v_0, v_{n+1}\}$,
$\EE' = \{(i, j, -d_{ij}) \mid (i,j,d_{ij})\in \EE\}\cup \{(v_0, i, 0), (i,
v_{n+1}, -p_i) \mid i\in \VV\}$, $v_0$ is the source node, and $v_{n+1}$ is the
sink node.
The digraph is referred as the activity-on-node network in the
literature (e.g.~\cite{Bartusch:etal:88,Neumann:Schwindt:97}).
If the digraph contains a negative-weight cycle then the \rcpspmax{} instance is
infeasible.
Otherwise the shortest path from the source $v_0$ to an activity~$i$ determines
the earliest possible start time for $i$, i.e. $-w(v_0\to i)$ where $w(.)$ is
the length of the path and the shortest path from an activity~$i$ to the
sink~$v_{n+1}$ the latest possible start time for~$i$ in any schedule, i.e.
$t_{max} + w(i\to v_{n+1})$.
The Bellman-Ford algorithm has a runtime complexity of $\cO(|\VV|\times |\EE|)$.

These earliest and latest start times can not only used for an initial smaller
domain, but also to improve the objective constraints by replacing them with
\begin{quote}\tt\small
\begin{tabbing}
mmmm\=mmmm\=mmmm\=mmmm\=\kill
\> \% Objective constraints\\
constraint\\
\>   forall (i in V) (S[i] + tail[i] <= objective);\\
\end{tabbing}
\end{quote}
where \texttt{tail[i]} is the ``negative'' length~$-w(i\to v_{n+1})$ of the
shortest path from~$i$ to $v_{n+1}$ in the digraph~$G$.
Preliminaries experiments confirmed that this modification
gave major improvements 
for solving an instance and generating a first solution, especially on
larger instances.
Another advantage specific to 
LCG is that a smaller initial domain also
reduces the size of the problem because less Boolean variables are necessary to
represent the integer domain in the SAT solver.

\subsection{Activities in Disjunction}
\label{ssec:actdisj}

Two activities $i$ and $j\in \VV$ are in \emph{disjunction}, if they cannot be
executed at the same time, i.e. their resource requirement for at least one
resource~$k\in\RR$ is bigger than the available capacity:
$r_{ik} + r_{jk} > R_k$.
Activities in disjunction can be exploited in order to reduce the search space.

The simplest way to model two activities~$i$ and~$j$ in disjunction is by two
propositional constraints sharing the same Boolean variable~$B_{ij}$.
\begin{align}
   \label{eq:holla}
   B_{ij} \to S_i + p_i \leq S_j && \forall i<j \text{ in disjunction}\\
   \neg B_{ij} \to S_j + p_j \leq S_i && \forall i<j \text{ in disjunction}
\end{align}
If $B_{ij}$ is $\true$ then $i$ must end before $j$ starts (denoted by
$i\before j$), and if $B_{ij}$ is $\false$ then $j\before i$.
The literals $B_{ij}$ and $\neg B_{ij}$ can be directly represented in the SAT
solver, consequently $B_{ij}$ represents the relation (structure) between these
activities.
The propagator of such a propositional constraint can only infer new bounds on
left hand side of the implication if the right hand side is false, and on the
start times variables if the left hand side is true.
For example, the right hand side in the second constraint is false if and only
if $\max_D{S_i} - \min_D{S_j} < p_j$. 
In this case the literal $\neg B_{ij}$ must be
false and therefore $i\before j$.

Adding these redundant constraints to the model allows the propagation
solver to more quickly determine information about start time variables.
The Zinc model of these constraints is
\begin{quote}\tt\small
\begin{tabbing}
mmmm\=mmmm\=mmmm\=mmmm\=\kill
\> \% Redundant non-overlapping (disjunctive) constraints\\
constraint\\
\> forall (i, j in V where i < j) (\\
\> \> if exists(res in R)(r[i, res] + r[j, res] > rcap[res]) then\\
\> \> \> \% Activity i must be run before or after j\\
\> \> \> let \{var bool: b\} in (\\
\> \> \> \> (b -> S[i] + p[i] <= S[j])\\
\> \> \> \> /\char`\\ \ (not(b) -> S[j] + p[j] <= S[i])\\
\> \> \> )\\
\> \> else true endif\\
\> );\\
\end{tabbing}
\end{quote}

The detection which activity runs before the other can be further improved by considering the domains of the start times, and the minimal distances in the activity-on-node-network (see e.g. Dorndorf et al.~\cite{Dorndorf:etal:00}).

\ignore{
Let the activities $i$ and $j$ be (tightly) bound by a minimal
$(i, j, d_{ij}^+)\in\EE$ and maximal $(j, i, d_{ji}^-)\in\EE$ time lag together,
so that $j$ always has a compulsory part during $i$'s execution, i.e. if
$-d_{ji}^- < p_i$ and $-d_{ji}^- < d_{ij}^+ + p_j$ hold.
If $o\in\VV$, and it exists a resource $k\in\RR$ with
$r_{ik} + r_{jk} + r_{ok} > R_k$ then
$o\before j \vee j\before o \vee i\before o$.
Note that no information can be retrieved if the activity~$o$ is with either~$i$
or ~$j$ in disjunction.

In the case if $o$ is in disjunction with both then model for activities in
disjunction~(\ref{eq:holla}) for the triple can be refined by
\begin{align*}
   B_{oi} \to S_o + p_o &\leq S_i \\
   \neg B_{oi} \to S_i + p_i &\leq S_o \\
   \neg B_{oi} \to S_j + p_j &\leq S_o\enspace,
\end{align*}
i.e. one constraint and the Boolean $B_{oj}$ can be omitted.
In the case if $o$ is not in disjunction with neither one then following
constraints can model this situation.
\begin{align*}
   B_{oj} \to S_o + p_o \leq S_j \\
   \neg B_{oj} \vee B_{io} \vee B_{jo}\\
   \neg B_{io} \leftrightarrow S_i + p_i \leq S_o \\
   \neg B_{jo} \leftrightarrow S_j + p_j \leq S_o
\end{align*}
This model is complicated and is questionable if it would help at all. The
problem here is that we do not know if $i$ or $j$ determines their overlapping
period. The known cases are if $d_{ij}^+ + p_j > p_i$ then $i$ determines the
end and if $-d_{ji}^- + p_j < p_i$ then $j$ determines the end. In both cases
the model can be simplified by using just one Boolean variable.
}

\section{The Branch-and-Bound Algorithm}
\label{sec:search}

Our branch-and-bound algorithms are based on deterministic and conflict-
driven branching strategies.
We use them solely or in combination as a hybrid where at first the
deterministic and then the conflict-driven branching is chosen (cf. Schutt et
al.~\cite{Schutt:etal:09}).
After each branch all constraints are propagated until a fixpoint is reached or
the inconsistency for the partial schedule or the instance is proven.
In the first case a new node is explored and in the second case an unexplored
branch is chosen if one exists or backtracking is performed.

\subsection{Deterministic Branching}

The deterministic branching strategy selects an unfixed start time variable
$S_i$ with the smallest possible start time~$\min_D{S_i}$. If there is a tie
between several variables then the variable with the biggest size, i.e.
$\max_D{S_i}-\min_D{S_i}$, is chosen.
If there is still a tie then the variable with the lowest index~$i$ is selected.
The binary branching is as follows: left branch $S_i \leq \min_D{S_i}$, and right
branch $S_i > \min_D{S_i}$.
We denote this branching strategy by \mslf{}.

This branching creates a time-oriented branch-and-bound algorithm similar to
Dorndorf et al.~\cite{Dorndorf:etal:00}, but it is simpler and does not involves
any dominance rule. Hence, it is weaker than their algorithm.

\subsection{Conflict-driven Branching}

The conflict-driven branching is a binary branching over literals in the SAT
solver. In the left branch the literal is set to $\true$ and in the right branch
to $\false$.
As described in Sec.~\ref{ssec:varrepr} on page~\pageref{ssec:varrepr} the
Boolean variables in the SAT solver represent values in the integer domain of a
variable~$x$ (e.g. $\neg \lit{x\le 3}$ ($\lit{x\le 10}$)) or a disjunction
between activities.
Hence, it creates a branch-and-bound algorithm that can be considered as a
mixture of time oriented and conflict-set oriented.

As branching heuristic the Variable-State-Independent-Decaying-Sum 
(\vsids)~\cite{Moskewicz:etal:01} is
used which is a part of the SAT solver.
In each branch it selects the literal with the highest activity counter where an
\emph{activity counter} is assigned to each literal, and is increased during
conflict analysis if the literal is related to the conflict.
The analysis results in a nogood which is added to the clause data base. Here,
we use the 1UIP as a nogood.

In order to accelerate the solution finding and increase the robustness of
the search on hard instances \vsids{} can be combined with restarts which
has been shown beneficial in SAT solving.  On restart the set of nogoods and
the activity counter has changed, so that the search will explore a very
different part of the search tree.  In the remainder \vsids{} with restart
is denoted by \restart{}.  Different restart policies can be applied, here a
geometric restart on nodes with an initial limit of 250 and a restart factor
of $2.0$ are used.

\subsection{Hybrid Branching}

At the beginning of each search the activity counters are 
all initialized with the
same value which can result in a poor performance of \vsids{}
at the start of search.
In order to avoid this situation at first \mslf{} can be chosen for branching
and then \vsids{} used 
after a restart is performed (e.g. after a specific number of
explored nodes).
This has the advantage that the deterministic search initializes the
activity counters with more meaningful values that can be fully exploited by
\vsids{}.
Here, we switch the searches after 
exploration of the first 500 nodes unless otherwise stated.
In the remainder we refer to the strategy as \HS{}.
Once more, the \vsids{} search after the first restart can benefit from
restart. We denote the hybrid branching approach with restarts by \HR{}.

\section{Computational Results}
\label{sec:exp}

We carried out experiments on \rcpspmax{} instances available 
from~\cite{rcpspmax:lib} and accessible from the PSPLib~\cite{psplib}. 
Our approach is compared to the best
known exact and non-exact methods so far on each testset.
At the website \texttt{http://www.cs.mu.oz.au/\~{}pjs/rcpsp} detailed results
can be obtained.

Our methods are evaluated on the following testsets which were systematically created using the instance generator ProGen/max (Schwindt~\cite{Schwindt:95}):
\begin{itemize}
\item \textbf{CD} ---
   1080 instances with 100 activities and 5 resources 
   (cf. Schwindt~\cite{Schwindt:98b}).
\item \textbf{UBO} ---
   \BCubo{10}, \BCubo{20}, \BCubo{50}, \BCubo{100}, and \BCubo{200}: each containing 90 instances with 5 resources and 10, 20, 50, 100, and 200 activities respectively 
   (cf. Franck et al.~\cite{Franck:etal:01}).
\item \textbf{SM} ---
   \BCj{10}, \BCj{20}, and \BCj{30}: each containing 270 instances with 5 resources and 10, 20, and 30 activities respectively 
   (cf. Kolisch et al.~\cite{Kolisch:etal:98}).
\end{itemize}
Note that although the 
testset SM consists of small instances they are considerably harder than e.g.
\BCubo{10} and \BCubo{20}. 

The experiments were run on Intel(R) Xeon(R) CPU E54052 processor with 2
GHz clock running GNU/Linux.
The code was written in Mercury
using the G12 Constraint Programming Platform and compiled with the Mercury
Compiler using grade hlc.gc.trseg. Each run was given a 10 minute runtime limit.

\subsection{Setup and Table Notations}

In order to solve each 
instance a two-phase process was used. 
Both phases used the basic model with the two extensions 
described in Subsections~\ref{ssec:initdomain} and~\ref{ssec:actdisj}.

In the first phase a \HS{} search was run to determine a first solution 
or to prove the infeasibility of the instance.
In contrast to the normal \HS{} we give the deterministic search 
more time to find a first solution and 
therefore we switch to \vsids{} only after after $5\times n$ nodes are 
explored, where $n$ is the number of activities.

The feasibility runs were set up with the trivial upper bound on the makespan
$t_{max} = \sum_{i\in\VV}\max(p_i,\max\{ d_{ij}\mid (i,j,d_{ij})\in \EE\})$.
The first phase was run until a solution (with makespan $UB$) was found
or infeasibility proved or the time limit reached.
In the first phase the the search strategy used should be at
both finding feasible solutions and proving infeasibility.
Hence, it could be exchanged with methods 
which might be more suitable than \HS{}.

In the second optimization phase, each feasible instance was set up
again this time with $t_{max} = \UB$. 
The tighter bound is highly beneficial to lazy clause generation since
it reduces the number of Boolean variables required to represent the problem.
The search for optimality was performed using one of the
various search strategies
defined in the previous section.

The execution of the two-phased process lead to the following measurements.
\begin{description}
\item[$rt_{max}$:] The runtime limit in seconds (for both phases together).
\item[$rt_{avg}$:] The average runtime in seconds (for both phases).
\item[fails:] The average number of fails perfomed in both phases of the
  search.
\item[\textsf{feas}:] The percentage of instances for which a solution was
found.
\item[\textsf{infeas}:] The percentage of instances for which the infeasibility
was proven.
\item[\textsf{opt}:] The percentage of instances for which an optimal solution
was found and proven.
\item[$\Delta_{LB}$:] The average distance from the best 
known lower bounds of feasible instances 
given in~\cite{rcpspmax:lib}.
\item[\textsf{\#svd}:] The number of instances which were proven to be
infeasible
or optimal.
\item[\textsf{cmpr(i)}:] Columns with this header give measurements only
related to those instances that were solved by each procedure where \textsf{i}
is the number of these instances.
\item[\textsf{all(i)}:] Columns with this header comare 
measurements for all instances examined in the experiment
where \textsf{i} is the number of these instances.
\end{description}

A note about special entries in the tables.  A table entry ``-'' indicates
no related number was available from previously published work.  A table
entry with two numbers the second in parentheses indicates the procedure was
applied several times: the first number is the average over all runs with
the second number, in parentheses, is the best number for all runs.  A table
entry marked ``$^\star$'' indicates the situation where a procedure was not
able to find a solution for all feasible instances and therefore the
corresponding number may not be comparable with the number for other
procedures in the same column.

\subsection{Comparison of the different strategies}

In the first experiment we compare all of our search strategies against each
other on all testsets.
The strategies are compared in terms of $rt_{avg}$ and failures for each test
set.

\begin{table}
   \centering
   \caption{Comparison on the testsets CD, UBO, and SM}
   \label{tab:comp:all}
\begin{tabular}{l|cr|rr|rr} 
\multirow{2}{*}{Procedure} & \multirow{2}{*}{\sf \#svd} & \multirow{2}{*}{$\Delta_{LB}$} & \multicolumn{2}{c|}{\sf cmpr(2230)} & \multicolumn{2}{c}{\sf all(2340)}\\ 
& & & $rt_{avg}$ & fails & $rt_{avg}$ & fails\\
\hline 
\mslf{} 
& 2237 & 3.96785 & 7.73 & 6804 & 35.96 & 23781\\ 
\mslf{} with restart 
& 2237 & 3.96352 & 7.80 & 6793 & 36.04 & 23787\\ 
\vsids{}
& 2276 & 3.76928 & 2.16 & 1567 & 22.91 & 13211\\ 
\restart{}
& 2276 & 3.73334 & \bf 2.02 & \bf 1363 & 22.38 & \bf 12212\\ 
\HS{}
& 2277 & 3.84003 & 2.22 & 1684 & 22.71 & 12933\\ 
\HR{}
& \bf 2278 & \bf 3.73049 & 2.04 & 1475 & \bf 22.36 & 12341\\ 
\end{tabular}
\end{table}

The results are summarized in the Table~\ref{tab:comp:all}. 
Similar to the
results for \rcpsp{} in Schutt et al.~\cite{Schutt:etal:10} 
all strategies using
\vsids{} are superior to the deterministic methods (\mslf{}), and similarly
competitive.
\HR{} is the most robust strategy, solving the most instances
to optimality and having the lowest $\Delta_{LB}$.
Restart is essential to make the search more robust
for the conflict-driven strategies, whereas the impact of restart 
on \mslf{} is minimal.

In contrast to the results in Schutt et al.~\cite{Schutt:etal:10} 
for \rcpsp{} the
conflict-driven searches were not uniformly superior to \mslf{}.
The three instances 67, 68, and 154 from \BCj{30} were
solved to optimality by \mslf{} and \mslf{} with restart, 
but neither \restart{} and \HR{}
could prove the optimality in the given time limit,
whereas \vsids{} and \HS{}
were not even able to find an optimal solution within the time limit.
Furthermore, our method could not find a first solution for the \BCubo{200}
instances 2, 4, and 70 nor prove the infeasibility for the \BCubo{200}
instance 40 within 10 minutes.
\ignore{
 For these instances 
we Only for these instance we let run our method
until a first solution was found or the infeasibility proven. The
corresponding number are included in the Table~\ref{tab:comp:all}. The
details about them follow later in the Subsection~\ref{ssec:resubo}.
}

\subsection{Results on the testset CD}

Table~\ref{tab:res:cd} presents the results for the testsets CD where
$98.1$\% ($1.9$\%) of the instances are feasible (infeasible).
Here, we compare \restart{} and \HR{} with the time-oriented branch-and-bound
procedure (\bbdorn{}) from Dorndorf et al.~\cite{Dorndorf:etal:00} and the evolutionary algorithm \evaball{} from 
Ballest{\'i}n et al.~\cite{Ballestin:etal:09}. 
The method \bbdorn{} performs better on this testset than the methods proposed
by De Reyck and Herroelen~\cite{DeReyck:Herroelen:98}, Schwindt~\cite{
Schwindt:98}\footnote{As reported in~\cite{Dorndorf:etal:00}}, and Fest et
al.~\cite{Fest:etal:99}. Moreover, \bbdorn{} is the best published exact method on this testset so far.
The \bbdorn{} method was implemented in C++ using \textsc{Ilog Solver}
and \textsc{Ilog Scheduler}. Their experiments were run on a Pentium Pro/200 PC
with NT 4.0 as operating system, thus their results were obtained on a machine
approximately ten times slower.

\begin{table}
   \centering
   \caption{Results on the testset CD}
   \label{tab:res:cd}
   \begin{threeparttable}
\begin{tabular}{lcccccc}
Procedure & $rt_{max}$ &  $rt_{avg}$ & \sf feas & \sf opt & \sf infeas & $\Delta_{LB}$\\
\hline

\bbdorn{} 
& 100 & - & \bf 98.1 & 71.7 & \bf 1.9 & $4.6$\tnote{a}\\
\ignore{
Smith and Pyle~\cite{Smith:Pyle:04}
& 10 & 1.85 & 98.1 & 63.0 & - & 6.8\\}
\evaball{}
& - & 0.62 & 98.1 & $\geq 65.9$ & - & \bf 3.24 (3.16)\\
\hline
\multirow{4}{*}{\restart{}}
& 1   & 0.38 & 97.9 & \bf 78.1 & 1.6 & 4.73$^\star$\\
& 10  & 1.39 & 98.1 & \bf 89.8 & 1.9 & \bf 3.20\\
& 100 & 6.17 & 98.1 & \bf 94.0 & 1.9 & \bf 2.86\\
& 600 & 19.32 & 98.1 & 95.8 & 1.9 & 2.81\\
\hline
\multirow{4}{*}{\HR{}}
& 1   & 0.44 & 97.9 & 76.8 & 1.6 & 4.87$^\star$\\
& 10  & 1.49 & 98.1 & 89.6 & 1.9 & \bf 3.20\\
& 100 & 6.27 & 98.1 & 93.9 & 1.9 & \bf 2.86\\
& 600 & 19.42 & 98.1 & \bf 96.0 & 1.9 & \bf 2.79
\end{tabular}

      \begin{tablenotes}
         \item[a] \footnotesize $\Delta_{LB}$ is based on the lower bounds presented in Schwindt~\cite{Schwindt:98b} which were not accessible for us.
      \end{tablenotes}
   \end{threeparttable}
\end{table}

We compare our results achieved with a runtime limit of 1 second to their
results with a limit of 100 seconds which should be clearly in favour of them.
While \bbdorn{} 
can prove feasibility and infeasibility of all instances, 
the first-phase \HS{} search with one second
was unable to prove infeasibilty of four infeasible instances
or find solutions to two feasible instances.
It does prove infeasibility of these four infeasible 
instances in less than $2.1$
seconds and finds a first solution for these two feasible instances 
in $4.8$ seconds and $5.04$ seconds respectively.
Within one second both our methods \restart{}
and \HR{} were able to prove the optimality of substantially 
more instances than \bbdorn{}.
With more time our methods are able to prove optimality of almost all
instances in these testsets.

One reason for the first-phase results at one second may simply be
that there is a
reasonable set up time required for lazy clause generation to generate all
the Boolean variables and hence there is not much time left for search.
Another reason for the weakness of proving infeasibility is that our model only
contains propagators that determine the order of activities in disjunction
concerning their domains, but not also their minimal distance in the transitive
closure of all precedences.\footnote{The missing propagators are not available
in the G12 Constraint Programming Platform.}
Dorndorf et al.~\cite{Dorndorf:etal:00} shows that these propagators are crucial
for detecting infeasibility.
That \HS{} is not so good at finding a first solution is not surprising, since
the search is not very problem specific as \bbdorn{}.
In order to overcome these problems one could run at first e.g. \bbdorn{} to
prove infeasibility and generate a first solution, and then apply our methods.

The method \evaball{} is the best published local search procedure on this
testset. Their results were obtained on a Samsung X15 Plus computer with
Pentium M processor with 1400 MHz clock speed. This means that our machine
is about 1.46 times faster than their.  Their limits are a maximum of 5000
schedules and a stop of the process if within 10 generation the best
schedule could not be improved.  Our methods generates better schedules
within 10 seconds than there approach, visible in the lower $\Delta_{LB}$ 
of 3.20.

Overall our methods are able to close 310 open problems and improve the
upper bound for all 21 remaining open problems in testset CD, according to
the results recorded in~\cite{rcpspmax:lib}.

\subsection{Results on testset \textsc{UBO}}
\label{ssec:resubo}

Table~\ref{tab:res:ubosmall} compares our procedures \restart{} and \HR{} 
with the truncated branch-and-bound methods \fbsfel{}, the heuristic~
\dmfel{}, and the genetic algorithm~\gafel{} all 
proposed by Franck et al.~\cite{Franck:etal:01}
on the UBO testset where $81.7$\% ($18.3$\%) of the instances are feasible 
(infeasible). 
In this table we add the column \textsf{feas + infeas} showing the sum of percentage of \textsf{feas} and \textsf{infeas} because the corresponding numbers for \fbsfel{} are not available.
Their results were obtained on personal computer PII with a 333MHz processor
running NT 4.0 as operating system, i.e. our machine is about 6.2 times faster.
They imposed a time limit of $n$ seconds, e.g. an instance with 100 activities
was given at most 100 seconds.
We compare our methods with 10 (100) times lower time limit which should 
be favorable to the other methods.

\begin{table}
   \centering
   \caption{Results on the testset UBO for \BCubo{10}, \BCubo{20}, 
   \BCubo{50} and \BCubo{100} instances in comparison with Franck et al.}
   \label{tab:res:ubosmall}
\begin{tabular}{lccccccc}
Procedure & $rt_{max}$ &  $rt_{avg}$ & \sf feas + infeas & feas & \sf opt & \sf infeas & $\Delta_{LB}$\\
\hline
\fbsfel{}
& $n$ & 12.4 & 99.66 & - & - & - & $6.82^\star$\\
\dmfel{}
& $n$ & 0.03 & 100 & 81.7 & - & 18.3 & $10.72$\\
\gafel{}
& $n$ & 3.16 & 100 & 81.7 & - & 18.3 & $6.93$\\
\hline
\multirow{2}{*}{\restart{}}
& $n/100$ & 0.21 & 95.0 & 80.0 & 70.8 & 15.0 & 5.73$^\star$\\
& $n/10$  & 0.78 & 100 & 81.7 & 75.3 & 18.3 & 4.99\\
\hline
\multirow{2}{*}{\HR{}}
& $n/100$ & 0.25 & 95.0 & 80.0 & 69.7 & 15.0 & 5.73$^\star$\\
& $n/10$  & 0.81 & 100 & 81.7 & 75.3 & 18.3 & 5.04\\
\end{tabular}
\end{table}

Their methods were able to prove the feasibility or infeasibility for all
instances (except one instance for the method \fbsfel{}).
Indeed \dmfel{} is extremely fast requiring just 0.03 seconds on
average, but it does not necessarily find very good solutions, as
shown by the high $\Delta_{LB}$. 

In contrast our first-phase was not always able to find a first solution
or prove infeasibilty with the time limit $n/100$.
No solution was found for 6 instances with 100 activities and the infeasibility
was not shown for 11 (1) instances with 100 (50) activities.
Once the time limit was extended to $n/10$ then the first phase was
always able to find a solution or prove infeasibility.
If we compare $\Delta_{LB}$ achieved with a time limit $n/10$ (note for a time
limit $n/100$ the data is not comparable, since our methods could not find a
solution for all feasible instances) then our methods have a substantially
better $\Delta_{LB}$ than their approaches, i.e., 
our methods are quicker in improving the makespan.
Our approaches could prove optimality for a substantial fraction of these
problems even with time limit $n/100$.

\begin{table}
   \centering
   \caption{Results on the testset UBO for \BCubo{10}, \BCubo{20}, 
   \BCubo{50} and \BCubo{100} instances in comparison with Ballest{\'i}n et al.}
   \label{tab:res:ubosmall:new}
\begin{tabular}{lcccccc}
Procedure & $rt_{max}$ &  $rt_{avg}$ & \sf feas & \sf opt & \sf infeas & $\Delta_{LB}$\\
\hline

\evaball{}
& - & 0.38 & 81.7 & - & - & \bf 4.82 (4.79)\\
\hline
\multirow{4}{*}{\restart{}}
& 1   & 0.22 & 80.0 & \bf 71.4 & 15.3 & 5.60$^\star$\\
& 10  & 0.89 & 81.7 & \bf 75.3 & 18.3 & 4.92\\
& 100 & 5.32 & 81.7 & \bf 77.2 & 18.3 & \bf 4.51\\
& 600 & 24.47 & 81.7 & \bf 78.1 & 18.3 & \bf 4.40\\
\hline
\multirow{4}{*}{\HR{}}
& 1   & 0.26 & 80.0 & 70.6 & 15.3 & 5.65$^\star$\\
& 10  & 0.92 & 81.7 & \bf 75.3 & 18.3 & 5.01\\
& 100 & 5.26 & 81.7 & \bf 77.2 & 18.3 & 4.55\\
& 600 & 24.14 & 81.7 & \bf 78.1 & 18.3 & 4.43
\end{tabular}

\end{table}

In the Table~\ref{tab:res:ubosmall:new} we compare our results with the best
local search method \evaball{} from Ballest{\'i}n et
al.~\cite{Ballestin:etal:09} on the UBO instances with at most 100 activities.
Their limits are a maximum of 5000 schedules and a stop of the evolution process after 10 generations if no better schedule could be found.
Our methods creates better schedules within 100 seconds than the
evolutionary algorithm \evaball{} leading to a smaller lower bound deviation.

\begin{table}
   \centering
   \caption{Results on the testset UBO for \BCubo{200} instances}
   \label{tab:res:ubo200}
   \begin{threeparttable}
\begin{tabular}{lccccccc}
Procedure & $rt_{max}$ &  $rt_{avg}$ & \sf feas & \sf opt & \sf infeas & $\Delta_{LB}$ & $\Delta_{\UB}$\\
\hline

\ifs{} 
& - & 2148.7 & 88.9 & - & - & - & 2.06\\
\ifsfr{} 
& - & 2024.7 & 88.9 & - & - & - & 1.81\\
\ifsmcsr{} 
& - & 1716.7 & 88.9 & - & - & - & 1.65\\
\hline
\multirow{2}{*}{\restart{}}
& 100 & 29.55 & 81.1 & 67.8 & 7.8 &  7.37$^\star$ & -0.41$^\star$\\
& 600 & 139.0 & 85.6 & \bf 68.9 & 10.0 & 10.11$^\star$ & -1.110$^\star$\\
& 600+\tnote{a} & 187.5 & 88.9 & \bf 68.9 & 11.1 & 11.88 & -1.249\\
\hline
\multirow{2}{*}{\HR{}}
& 100 & 29.9 & 81.1 & \bf 68.9 & 7.8 & \bf 7.22$^\star$ & \bf -0.48$^\star$\\
& 600 & 139.0 & 85.6 & \bf 68.9 & 10.0 & \bf 10.10$^\star$ & \bf -1.111$^\star$\\
& 600+\tnote{a} & 186.9 & 88.9 & \bf 68.9 & 11.1 & \bf 11.87 & \bf -1.250
\end{tabular}
      \begin{tablenotes}
         \item[a] \footnotesize For comparison purpose the instances 2, 4, 40, and 70 were run until a first solution was found or infeasibility proven.
      \end{tablenotes}
   \end{threeparttable}
\end{table}

The Table~\ref{tab:res:ubo200} presents the 
results on \BCubo{200} which are compared to the iterative flattening searches \ifs{}, \ifsfr{}, and \ifsmcsr{} from Oddi and Rasconi~\cite{Oddi:Rasconi:09}.\footnote{No machine details are given in \cite{Oddi:Rasconi:09}.}
The table contains the extra column $\Delta_{\UB}$ that reports the average distance from the best known upper bounds of feasible instances given 
in~\cite{rcpspmax:lib}.
\ignore{Their local search method is an iterative flattening search that improves schedules by escape strategies for stall situations.}
Note Franck et al.~\cite{Franck:etal:01} also run their methods on \BCubo{200}, but the presented results are accumulated with the results on instances with 500 and 1000 activities, so that a comparison is not possible.

Within the given time limit $\HS{}$ was not able to find a solution for the instances 2, 4, and 70 and to prove the infeasibility for the instance 40.
In order to compare the results with Oddi and Rasconi we let the first phase
of our solution method until a solution was found or infeasibility proven.
The runtimes for the instances 2, 4, 40, and 70 were 
1030, 1478, 1139, and 3103 seconds respectively.
Interestingly, the first solutions obtained by our method
for the instances 2, 4, and 70 have a better upper bound by 
62, 46, and 37 
respectively than the previously best known upper 
bound recorded in~\cite{rcpspmax:lib}.

Comparing these results on $\Delta_{\UB}$ with Oddi and Rasconi clearly our procedures achieve better schedules. \restart{} and \HR{} perform comparably.
The \BCubo{200} instances clearly show that \HS{} as the search strategy in
the first phase can have problems 
to find a first solution or to prove infeasibility.

In total our approaches close 178 open instances and improve the upper bound
for 27 instances of 31 remaining open instances with 200 activities or less
in the testset UBO, according to the results recorded
in~\cite{rcpspmax:lib}.

\subsection{Results on testset SM}

Finally for 
the testset SM we compare our approaches \mslf{}, \restart{}, and \HR{} with
method \bbs{} from Schwindt~\cite{Schwindt:98}\footnote{The paper~\cite{
Schwindt:98} was not accessible for us, so that here the reported results are
taken from~\cite{Cesta:etal:02}.}, \ises{} from Cesta et
al.~\cite{Cesta:etal:02}, and \swo{} from Smith and Pyle~\cite{Smith:Pyle:04}.
The method \bbs{}~\cite{Schwindt:98} is a branch-and-bound algorithm that resolves resource
conflicts by adding precedences between activities and has been run on a Pentium
200 with a 100 seconds time limit.
\ises{}~\cite{Cesta:etal:02} is a heuristic that also adds precedences between activities in order to
resolve/avoid resource conflicts, uses restarts and has been run on a SUN
UltraSparc 30 (266 MHz) with the same time limit.
The method \swo{}~\cite{Smith:Pyle:04} 
is a squeaky wheel optimisation. Their methods is divided into
two stages: schedule generation and prioritisation where the schedule is created
by a heuristic with priority scheme and the latter changes the priorities on
variables depending on how ``well'' it is handled in the former stage.
Their benchmarks were performed on a 1700 Mhz Pentium 4.
Note that \ises{} and \swo{} are not exact methods, i.e. they cannot prove
infeasibility unless the precedence graph contains a positive weight cycle, and
optimality is only proven if the makespan of the solution found 
equals the known lower bound.

\begin{table}
   \centering
   \caption{Results on the \BCj{30}}
   \label{tab:res:j30}
   \begin{threeparttable}
\begin{tabular}{lcccccc}
Procedure & $rt_{max}$ &  $rt_{avg}$ & \sf feas & \sf opt & \sf infeas & $\Delta_{LB}$\\
\hline

\bbs{} 
& 100 & - & 67.7 & 42.6 & - & $9.56$\tnote{a}\\
\ises{} 
& 100 & 22.68 & 68.5 & 33.9 (35.6) & - & $10.99$ ($10.37$)\\
\swo{}
& 10 & 1.07 & 68.5 & 35.0 & - & 10.3\\
\hline
\mslf{} 
& 1   & 0.16 & 68.5 & 58.1 & 31.5 & 8.91\\
& 10  & 0.82 & 68.5 & 61.9 & 31.5 & 8.40\\
& 100 & 4.90 & 68.5 & \bf 64.8 & 31.5 & 8.23\\
& 600 & 21.61 & 68.5 & \bf 65.5 & 31.5 & 8.20\\
\hline
\ignore{
\mslf{}(restart)
& 1   & 0.16 & 68.5 & 58.5 & 31.5 & 8.98\\
& 10  & 0.83 & 68.5 & 61.9 & 31.5 & 8.41\\
& 100 & 4.96 & 68.5 & 64.8 & 31.5 & 8.22\\
& 600 & 21.77 & 68.5 & 65.5 & 31.5 & 8.20\\
\hline
\vsids{}
& 1   & 0.11 & 68.5 & 61.9 & 31.5 & 8.41\\
& 10  & 0.58 & 68.5 & 64.4 & 31.5 & 8.27\\
& 100 & 3.91 & 68.5 & 64.8 & 31.5 & 8.21\\
& 600 & 21.16 & 68.5 & 65.2 & 31.5 & 8.19\\
\hline}
\restart{}
& 1   & 0.12 & 68.5 & \bf 61.5 & 31.5 & 8.38\\
& 10  & 0.57 & 68.5 & \bf 64.1 & 31.5 & 8.19\\
& 100 & 3.92 & 68.5 & \bf 64.8 & 31.5 & 8.17\\
& 600 & 21.34 & 68.5 & 65.2 & 31.5 & \bf 8.12\\
\hline
\ignore{
\HS{}
& 1   & 0.12 & 68.5 & 61.9 & 31.5 & 8.44\\
& 10  & 0.55 & 68.5 & 64.8 & 31.5 & 8.22\\
& 100 & 3.88 & 68.5 & 64.8 & 31.5 & 8.16\\
& 600 & 21.18 & 68.5 & 65.2 & 31.5 & 8.14\\
\hline}
\HR{}
& 1   & 0.12 & 68.5 & \bf 61.5 & 31.5 & \bf 8.37\\
& 10  & 0.59 & 68.5 & \bf 64.4 & 31.5 & \bf 8.18\\
& 100 & 3.93 & 68.5 & \bf 64.8 & 31.5 & \bf 8.16\\
& 600 & 21.47 & 68.5 & 65.2 & 31.5 & 8.13
\end{tabular}

      \begin{tablenotes}
         \item[a] \footnotesize $\Delta_{LB}$ is based on the lower bounds presented in Schwindt~\cite{Schwindt:98b} which were not accessible for us.
      \end{tablenotes}
   \end{threeparttable}
\end{table}

Table~\ref{tab:res:j30} presents the results for the 270 instances from SM 
with 30
activities. From these instances 185, i.e. $68.5\%$ are feasible and 85, i.e.
$31.5\%$ infeasible.
All our approaches could prove feasibility and infeasibility of all instances
within one second whereas \bbs{} could not find a solution for a few feasible
instances. Moreover, our methods could prove optimality significantly more
often than the exact method \bbs{} (and clearly also the incomplete methods).
All our methods were able to find on average better
solution in one seconds than these approaches as 
indicated by a lower $\Delta_{LB}$. 
For these harder benchmarks our methods clearly outperform the competition.
One reason could be
that constraint propagation over the \cumu{} constraint 
has a greater benefit
than on other testsets because here more activities can be run simultaneously.

Our approaches each give 
similar results: \restart{} and \HR{} are superior to
\mslf{} until 10 seconds,  and all are similar each other with longer time
limits.
For this problem set \mslf{} could be prove
optimality for three instances where \restart{} and \HR{} only found the
optimal solution.
On the other hand \mslf{} could not find an optimal
solution for two instance where \restart{} and \HR{} could.
It seems that \mslf{} may better suits problems where more activities can be
executed in parallel, but this needs further investigation.

Experiments were also carried out on the instances with 10 or 20
activities. 
All out methods could solve all 270 instances with 10 activities 
within 0.05 seconds.
And all out methods could solve all 270 instances with 20 activities 
within 30 seconds. 
Moreover for the instances with 20 activities an
optimal solution was found within 1 second for all feasible instances.

Here, our approaches close 85 open problems and improve the upper bound for 3 problems of the 6 remaining open problems in the testset SM, according to the results recorded in~\cite{rcpspmax:lib}.

\ignore{
The instance UBO200-4 has an upper bound of 935 and the activities start times
are 7 797 190 522  78   0 763 127 608 238  84   8  23 159 493 348 921 172 830
843 634 604 222 698 797 836 590 145  78 634 172  95 810 776 286 879 725 686 225
305 602 157 242 258 364 662 515  26 504 628 231 803  12 680 451 600 590 467 643
670 780 327 909 696 155 353 239 565 569 141 550 202 559 386  43 315 532 741 505
545 652 863 917 854 460 704 690 766 652 579 535 242 794  53 338 143 619 611 830
428 457 632  18 469 698 267 810 626 718 269 288 294 115 413 585 803 828 133 319
359 679 486  87 702 891 437 790 748 473  37 338 306  53 573 611 146  28 115 900
407 481 837 400 328 371 761 708 407 395 819 163 339 686 733 872 279 472 180 190
38 294 644 389 336 759 446  84  12 848 889 250 610 851 879  73  63 511 421 107
913 151 212 756 136 262 793 309 122 303 463 746 600  84 932 840 913 719 924 643
381. This solution was found after 17 hours 51 minutes and 40 seconds.}

\section{Conclusion}
\label{sec:conc}

In this paper we minimize the project duration of \rcpspmax{} 
using a generic 
constraint programming solver that includes nogood learning facilities and 
conflict-driven search.
Experiments on three well-established benchmark suites show that
our solver is able to find better solutions quicker than competing
approaches, and prove optimality for many more instances than competing
approaches. 

We use a two-phase process. In the first phase a solution is
generated or infeasibility is proven and in the second phase a branch-and-bound
algorithm is used for optimization where the problem is set up with an upper
bound on the project duration found from the first solution.
In contrast to some previous approaches 
we use individual propagators for precedence constraints
instead of propagators taking all precedences into account at once.
This yields not only to weaker propagation, but also slower detection of
infeasibility, in particular for instances with a large
number of precedences like for instances with 100 activities or more from the testset UBO.
Hence our generic search used in the first phase is
sometimes slower in finding a
first solution than other problem-specific approaches in the literature.
However, the first-phase generic search could 
be replaced by one of these methods.

Overall, our method could close 573 open problems and improve a further 51
upper bounds on the project duration from the 58 remaining open problems, according to the best known results given in~\cite{rcpspmax:lib}.
We note though that the methods from 
Ballest{\'i}n et al.~\cite{Ballestin:etal:09}, and Oddi and 
Rasconi~\cite{Oddi:Rasconi:09} may have found 
better upper bounds than are recorded in~\cite{rcpspmax:lib} 
on some problems, but we could not find any record of this.
\ignore{
We note though that previous methods (in particular~\cite{} \pjs{which?})
will have almost certainly closed some of these instances, but we can find
no record of the instances they closed nor the optimal solutions found.}
Note that our method is highly robust: 
our method proves the best known optimal for each already closed instance
in every testset. 
Furthermore, for every open instance in every testset
we either close the instance or improve the upper bounds,
except for 7 instances (and here we regenerate the 
best known upper bound for 4 of them).

\begin{acknowledgements}
National ICT Australia is funded by the Australian Government as represented
by the Department of Broadband,
Communications and the Digital Economy and the Australian Research Council.
\end{acknowledgements}

\bibliographystyle{acm}      
\bibliography{refs_rcpspmax}

\clearpage
\begin{WithAppendix}
\appendix

\section{Closed instances}
\label{sec:ClosedInsts}

In the Tables~\ref{tab:closed:c}--\ref{tab:closed:ubo200} all 573 
previously open instances
(regarding to the reported results in~\cite{rcpspmax:lib}) are listed that had been closed by one of our methods.
For each instance following parameters are given the instance number (Inst), the
previously best known upper bound (Best $\UB$) on the makespan, the proved optimal makespan
(Optimal) and the lowest runtime (Best $rt$) of our methods which could solve the instance till optimality.
Optimal makespan are written in italic if the makespan is lower than the
previously best known upper bound.
Note that some of these instances were presumably 
closed by other methods, but we can find no record of either the
instance number or the optimal value.

\setcounter{LTchunksize}{3}
\begin{center}
\begin{longtable}{rrrr|rrrr|rrrr}
\caption{All closed instances from class \BCc}
\label{tab:closed:c}\\
999 & 999 & 999 & 555.55& 999 & 999 & 999 & 555.55& 999 & 999 & 999 & 555.55
\kill
\begin{sideways}Inst\end{sideways} &
\begin{sideways}Best $\UB$\end{sideways} &
\begin{sideways}Optimal\end{sideways} &
\begin{sideways}Best $rt$\end{sideways}&
\begin{sideways}Inst\end{sideways} &
\begin{sideways}Best $\UB$\end{sideways} &
\begin{sideways}Optimal\end{sideways} &
\begin{sideways}Best $rt$\end{sideways}&
\begin{sideways}Inst\end{sideways} &
\begin{sideways}Best $\UB$\end{sideways} &
\begin{sideways}Optimal\end{sideways} &
\begin{sideways}Best $rt$\end{sideways}\\ \hline \hline
\endfirsthead
\hline
\multicolumn{12}{|l|}%
{\textbf{\tablename\ \thetable{} -- continued from previous page}}\\
\hline
999 & 999 & 999 & 555.55& 999 & 999 & 999 & 555.55& 999 & 999 & 999 & 555.55
\kill
\begin{sideways}Inst\end{sideways} &
\begin{sideways}Best $\UB$\end{sideways} &
\begin{sideways}Optimal\end{sideways} &
\begin{sideways}Best $rt$\end{sideways}&
\begin{sideways}Inst\end{sideways} &
\begin{sideways}Best $\UB$\end{sideways} &
\begin{sideways}Optimal\end{sideways} &
\begin{sideways}Best $rt$\end{sideways}&
\begin{sideways}Inst\end{sideways} &
\begin{sideways}Best $\UB$\end{sideways} &
\begin{sideways}Optimal\end{sideways} &
\begin{sideways}Best $rt$\end{sideways}\\ \hline \hline
\endhead
\hline
\multicolumn{12}{|l|}%
{Continued on next page}\\
\hline
\endfoot
\hline \hline
\endlastfoot
1 & 336 & 336 & 0.61 & 3 & 379 & 379 & 0.24 & 4 & 258 & 258 & 0.70\\
6 & 336 & \itshape 327 & 0.44 & 12 & 331 & 331 & 0.24 & 32 & 370 & \itshape 367
& 6.12\\
33 & 383 & 383 & 2.43 & 34 & 421 & \itshape 391 & 60.40 & 35 & 259 & \itshape
254 & 221.81\\
37 & 325 & \itshape 312 & 129.92 & 38 & 306 & \itshape 291 & 40.96 & 39 & 428 &
\itshape 421 & 5.10\\
40 & 395 & \itshape 386 & 1.45 & 62 & 621 & \itshape 602 & 10.72 & 64 & 688 &
688 & 0.39\\
65 & 376 & \itshape 355 & 60.37 & 90 & 293 & 293 & 0.19 & 91 & 260 & 260 & 0.74
\\
92 & 360 & 360 & 0.24 & 94 & 428 & 428 & 0.22 & 95 & 399 & 399 & 0.58\\
96 & 501 & 501 & 0.81 & 97 & 489 & 489 & 0.53 & 98 & 518 & 518 & 0.55\\
100 & 399 & 399 & 0.42 & 121 & 410 & \itshape 399 & 6.19 & 124 & 304 & \itshape
270 & 26.90\\
126 & 502 & 502 & 2.64 & 127 & 404 & \itshape 401 & 0.24 & 128 & 505 & 505 &
0.53\\
129 & 517 & \itshape 506 & 9.26 & 130 & 434 & \itshape 417 & 28.87 & 153 & 554 &
\itshape 553 & 3.76\\
154 & 535 & \itshape 524 & 12.41 & 155 & 375 & \itshape 361 & 201.56 & 156 & 399
& \itshape 387 & 134.98\\
157 & 475 & \itshape 453 & 1.89 & 158 & 397 & 397 & 1.28 & 159 & 488 & 488 &
2.44\\
165 & 371 & 371 & 3.17 & 181 & 456 & 456 & 0.36 & 182 & 376 & 376 & 0.37\\
183 & 461 & 461 & 0.45 & 185 & 370 & \itshape 364 & 2.36 & 186 & 410 & 410 &
0.40\\
188 & 321 & \itshape 307 & 0.97 & 190 & 401 & 401 & 0.19 & 191 & 493 & 493 &
0.13\\
211 & 445 & 445 & 0.88 & 212 & 564 & 564 & 0.89 & 213 & 710 & 710 & 1.01\\
214 & 624 & 624 & 1.56 & 217 & 365 & \itshape 362 & 2.64 & 220 & 403 & \itshape
393 & 1.95\\
224 & 304 & 304 & 0.45 & 242 & 431 & \itshape 425 & 76.30 & 243 & 533 & \itshape
519 & 7.76\\
244 & 514 & \itshape 508 & 4.74 & 246 & 574 & 574 & 1.54 & 247 & 478 & \itshape
471 & 15.98\\
248 & 443 & \itshape 430 & 31.29 & 249 & 635 & \itshape 633 & 1.95 & 251 & 308 &
308 & 1.21\\
260 & 469 & 469 & 0.54 & 271 & 498 & \itshape 497 & 2.84 & 272 & 277 & 277 &
0.39\\
273 & 598 & \itshape 579 & 1.05 & 277 & 448 & \itshape 410 & 2.08 & 278 & 587 &
587 & 0.27\\
280 & 451 & 451 & 0.30 & 301 & 412 & 412 & 1.33 & 303 & 329 & \itshape 319 &
1.53\\
304 & 346 & \itshape 333 & 132.18 & 306 & 296 & \itshape 288 & 1.00 & 307 & 342
& \itshape 309 & 110.16\\
308 & 564 & \itshape 545 & 19.74 & 309 & 503 & 503 & 2.87 & 314 & 329 & \itshape
324 & 2.03\\
315 & 294 & 294 & 0.49 & 328 & 255 & 255 & 0.34 & 332 & 336 & \itshape 326 &
15.10\\
333 & 410 & \itshape 404 & 3.71 & 335 & 426 & \itshape 413 & 3.03 & 338 & 415 &
\itshape 403 & 145.06\\
340 & 322 & \itshape 312 & 12.46 & 346 & 446 & 446 & 8.43 & 349 & 451 & \itshape
444 & 17.53\\
361 & 523 & \itshape 513 & 5.18 & 363 & 566 & 566 & 1.77 & 364 & 372 & \itshape
360 & 0.36\\
365 & 445 & 445 & 0.65 & 366 & 419 & 419 & 0.25 & 367 & 322 & 322 & 0.32\\
369 & 390 & 390 & 0.21 & 391 & 323 & \itshape 314 & 6.78 & 392 & 322 & \itshape
311 & 30.72\\
393 & 337 & \itshape 331 & 0.37 & 394 & 469 & 469 & 0.50 & 397 & 588 & \itshape
524 & 8.76\\
399 & 315 & \itshape 290 & 72.91 & 400 & 420 & \itshape 411 & 2.21 & 406 & 362 &
362 & 0.22\\
413 & 344 & 344 & 0.36 & 421 & 469 & \itshape 458 & 7.04 & 422 & 794 & \itshape
776 & 33.42\\
423 & 401 & 401 & 1.11 & 424 & 394 & \itshape 382 & 74.41 & 426 & 350 & \itshape
333 & 302.85\\
427 & 314 & \itshape 308 & 2.07 & 428 & 831 & 831 & 30.62 & 430 & 361 & \itshape
345 & 108.07\\
433 & 372 & \itshape 369 & 9.68 & 435 & 361 & \itshape 359 & 3.53 & 440 & 260 &
\itshape 258 & 5.52\\
451 & 365 & 365 & 0.25 & 452 & 420 & \itshape 419 & 0.62 & 453 & 659 & 659 &
5.45\\
454 & 498 & \itshape 493 & 0.65 & 455 & 304 & 304 & 0.29 & 456 & 609 & 609 &
0.26\\
457 & 430 & \itshape 428 & 0.29 & 458 & 402 & 402 & 0.24 & 459 & 499 & \itshape
447 & 1.26\\
481 & 433 & \itshape 420 & 1.22 & 482 & 905 & 905 & 16.12 & 483 & 426 & \itshape
402 & 4.73\\
484 & 574 & 574 & 0.58 & 486 & 586 & \itshape 568 & 15.16 & 487 & 734 & 734 &
11.47\\
488 & 485 & \itshape 483 & 0.46 & 489 & 397 & \itshape 382 & 6.55 & 490 & 462 &
462 & 0.22\\
493 & 503 & 503 & 0.21 & 495 & 353 & 353 & 0.16 & 497 & 333 & \itshape 323 &
2.58\\
511 & 440 & 440 & 1.32 & 513 & 555 & \itshape 551 & 0.80 & 514 & 501 & \itshape
489 & 3.21\\
515 & 715 & \itshape 673 & 19.99 & 516 & 394 & \itshape 393 & 1.17 & 517 & 407 &
\itshape 399 & 0.78\\
518 & 424 & \itshape 418 & 8.86 & 519 & 437 & 437 & 0.78 & 520 & 567 & \itshape
560 & 1.38\\
523 & 389 & 389 & 1.86 & 530 & 292 & 292 & 0.17 & 538 & 308 & 308 & 0.38\\
540 & 310 & 310 & 8.64 & & & & & & & &\\
\end{longtable}
\begin{longtable}{rrrr|rrrr|rrrr}
\caption{All closed instances from class \BCd}
\label{tab:closed:d}\\
999 & 999 & 999 & 555.55& 999 & 999 & 999 & 555.55& 999 & 999 & 999 & 555.55
\kill
\begin{sideways}Inst\end{sideways} &
\begin{sideways}Best $\UB$\end{sideways} &
\begin{sideways}Optimal\end{sideways} &
\begin{sideways}Best $rt$\end{sideways}&
\begin{sideways}Inst\end{sideways} &
\begin{sideways}Best $\UB$\end{sideways} &
\begin{sideways}Optimal\end{sideways} &
\begin{sideways}Best $rt$\end{sideways}&
\begin{sideways}Inst\end{sideways} &
\begin{sideways}Best $\UB$\end{sideways} &
\begin{sideways}Optimal\end{sideways} &
\begin{sideways}Best $rt$\end{sideways}\\ \hline \hline
\endfirsthead
\hline
\multicolumn{12}{|l|}%
{\textbf{\tablename\ \thetable{} -- continued from previous page}}\\
\hline
999 & 999 & 999 & 555.55& 999 & 999 & 999 & 555.55& 999 & 999 & 999 & 555.55
\kill
\begin{sideways}Inst\end{sideways} &
\begin{sideways}Best $\UB$\end{sideways} &
\begin{sideways}Optimal\end{sideways} &
\begin{sideways}Best $rt$\end{sideways}&
\begin{sideways}Inst\end{sideways} &
\begin{sideways}Best $\UB$\end{sideways} &
\begin{sideways}Optimal\end{sideways} &
\begin{sideways}Best $rt$\end{sideways}&
\begin{sideways}Inst\end{sideways} &
\begin{sideways}Best $\UB$\end{sideways} &
\begin{sideways}Optimal\end{sideways} &
\begin{sideways}Best $rt$\end{sideways}\\ \hline \hline
\endhead
\hline
\multicolumn{12}{|l|}%
{Continued on next page}\\
\hline
\endfoot
\hline \hline
\endlastfoot
2 & 488 & 488 & 0.71 & 3 & 359 & \itshape 351 & 1.22 & 6 & 483 & \itshape 475 &
1.90\\
7 & 371 & 371 & 0.47 & 9 & 558 & \itshape 552 & 7.16 & 10 & 430 & \itshape 428 &
1.89\\
11 & 400 & 400 & 0.13 & 31 & 581 & \itshape 562 & 35.64 & 32 & 603 & \itshape
588 & 285.56\\
33 & 448 & \itshape 445 & 46.95 & 34 & 489 & \itshape 466 & 5.93 & 36 & 674 &
674 & 11.08\\
37 & 529 & 529 & 0.67 & 38 & 496 & \itshape 490 & 2.00 & 40 & 491 & 491 & 0.80\\
61 & 476 & 476 & 2.02 & 62 & 717 & \itshape 710 & 9.69 & 64 & 611 & \itshape 596
& 6.12\\
65 & 539 & \itshape 493 & 138.61 & 66 & 472 & \itshape 449 & 323.21 & 67 & 501 &
\itshape 483 & 4.92\\
68 & 582 & \itshape 554 & 123.64 & 70 & 622 & \itshape 613 & 235.48 & 71 & 356 &
\itshape 354 & 2.34\\
88 & 394 & 394 & 1.12 & 91 & 502 & 502 & 0.41 & 92 & 407 & 407 & 0.28\\
93 & 392 & \itshape 387 & 1.09 & 94 & 457 & 457 & 1.13 & 96 & 450 & \itshape 445
& 0.22\\
97 & 468 & \itshape 464 & 2.49 & 98 & 540 & \itshape 527 & 0.85 & 100 & 447 &
447 & 0.49\\
122 & 665 & \itshape 656 & 4.19 & 123 & 497 & \itshape 481 & 4.63 & 124 & 634 &
\itshape 623 & 7.57\\
125 & 492 & \itshape 491 & 0.76 & 126 & 413 & \itshape 393 & 40.83 & 127 & 445 &
\itshape 432 & 0.95\\
128 & 661 & \itshape 626 & 43.29 & 130 & 623 & \itshape 616 & 13.47 & 134 & 455
& \itshape 454 & 0.76\\
151 & 575 & \itshape 560 & 7.24 & 152 & 467 & \itshape 456 & 1.25 & 153 & 535 &
\itshape 514 & 204.89\\
155 & 463 & \itshape 440 & 122.23 & 156 & 545 & \itshape 514 & 152.52 & 157 &
670 & \itshape 651 & 65.59\\
160 & 546 & \itshape 530 & 9.81 & 165 & 418 & 418 & 0.98 & 166 & 426 & \itshape
414 & 52.70\\
169 & 386 & 386 & 0.74 & 182 & 464 & 464 & 0.37 & 184 & 659 & 659 & 0.64\\
189 & 608 & 608 & 0.97 & 190 & 504 & 504 & 0.47 & 211 & 670 & \itshape 668 &
4.53\\
213 & 449 & \itshape 440 & 1.77 & 214 & 457 & \itshape 455 & 0.70 & 215 & 627 &
627 & 1.90\\
217 & 635 & 635 & 1.00 & 218 & 605 & 605 & 0.60 & 219 & 458 & \itshape 445 &
2.05\\
220 & 685 & \itshape 677 & 0.95 & 225 & 615 & 615 & 0.14 & 241 & 827 & \itshape
782 & 26.30\\
242 & 803 & \itshape 765 & 96.46 & 243 & 739 & \itshape 713 & 9.88 & 245 & 520 &
\itshape 507 & 3.56\\
247 & 481 & \itshape 443 & 104.31 & 249 & 707 & \itshape 684 & 31.53 & 251 & 484
& 484 & 0.24\\
252 & 600 & 600 & 0.35 & 254 & 402 & 402 & 5.17 & 258 & 543 & 543 & 0.88\\
272 & 597 & 597 & 0.51 & 274 & 474 & \itshape 473 & 0.48 & 275 & 613 & 613 &
1.07\\
276 & 526 & \itshape 523 & 0.46 & 277 & 575 & \itshape 569 & 0.41 & 280 & 665 &
665 & 0.71\\
303 & 485 & \itshape 473 & 2.33 & 304 & 451 & 451 & 2.84 & 305 & 644 & \itshape
634 & 3.56\\
306 & 596 & \itshape 578 & 1.46 & 309 & 761 & \itshape 754 & 2.18 & 332 & 699 &
\itshape 682 & 75.84\\
333 & 588 & \itshape 583 & 1.58 & 336 & 658 & 658 & 0.63 & 337 & 531 & \itshape
506 & 6.67\\
338 & 632 & 632 & 1.12 & 339 & 839 & 839 & 31.77 & 340 & 770 & \itshape 753 &
184.18\\
343 & 522 & 522 & 0.29 & 346 & 432 & 432 & 0.18 & 348 & 457 & 457 & 0.46\\
355 & 431 & 431 & 0.53 & 358 & 588 & 588 & 0.21 & 361 & 544 & 544 & 0.63\\
363 & 430 & 430 & 2.50 & 369 & 504 & 504 & 0.46 & 370 & 662 & 662 & 0.34\\
391 & 655 & \itshape 653 & 3.93 & 392 & 624 & 624 & 1.22 & 393 & 655 & \itshape
654 & 0.99\\
394 & 507 & \itshape 487 & 2.95 & 396 & 691 & \itshape 687 & 1.47 & 397 & 636 &
636 & 1.72\\
398 & 422 & \itshape 400 & 3.66 & 399 & 546 & \itshape 543 & 1.90 & 400 & 723 &
723 & 0.87\\
419 & 551 & 551 & 0.17 & 421 & 589 & \itshape 550 & 77.41 & 422 & 729 & 729 &
0.63\\
423 & 791 & \itshape 776 & 67.25 & 424 & 789 & \itshape 757 & 21.17 & 425 & 813
& \itshape 783 & 17.20\\
426 & 707 & 707 & 1.00 & 427 & 592 & 592 & 0.81 & 428 & 584 & 584 & 0.99\\
429 & 663 & \itshape 629 & 2.33 & 430 & 770 & \itshape 768 & 7.14 & 431 & 394 &
394 & 0.71\\
436 & 477 & 477 & 0.42 & 437 & 578 & 578 & 0.38 & 440 & 619 & 619 & 0.23\\
452 & 616 & \itshape 610 & 0.85 & 455 & 546 & 546 & 0.28 & 456 & 676 & 676 &
0.57\\
457 & 553 & 553 & 0.21 & 458 & 538 & 538 & 0.53 & 462 & 373 & 373 & 0.12\\
470 & 483 & 483 & 0.17 & 481 & 546 & 546 & 0.47 & 482 & 656 & 656 & 1.61\\
483 & 578 & 578 & 0.31 & 484 & 795 & 795 & 0.81 & 485 & 622 & \itshape 615 &
1.56\\
486 & 692 & 692 & 1.44 & 487 & 663 & 663 & 0.96 & 489 & 630 & \itshape 615 &
4.93\\
490 & 778 & 778 & 1.02 & 498 & 508 & 508 & 0.14 & 501 & 669 & 669 & 0.15\\
504 & 438 & 438 & 0.16 & 507 & 527 & 527 & 0.21 & 511 & 719 & 719 & 1.28\\
512 & 580 & \itshape 562 & 4.30 & 513 & 576 & \itshape 566 & 2.47 & 514 & 800 &
800 & 1.48\\
516 & 801 & 801 & 0.79 & 517 & 603 & \itshape 602 & 0.92 & 518 & 709 & 709 &
2.87\\
519 & 618 & \itshape 600 & 3.25 & 520 & 695 & 695 & 2.32 & 522 & 431 & 431 &
0.27\\
523 & 409 & 409 & 0.75 & 524 & 492 & \itshape 490 & 0.65 & 528 & 450 & 450 &
0.27\\
529 & 421 & 421 & 0.29 & 530 & 486 & 486 & 0.29 & 540 & 510 & 510 & 0.74\\
\end{longtable}
\begin{longtable}{rrrr|rrrr|rrrr}
\caption{All closed instances from class \BCj{20}}
\label{tab:closed:j20}\\
999 & 999 & 999 & 555.55& 999 & 999 & 999 & 555.55& 999 & 999 & 999 & 555.55
\kill
\begin{sideways}Inst\end{sideways} &
\begin{sideways}Best $\UB$\end{sideways} &
\begin{sideways}Optimal\end{sideways} &
\begin{sideways}Best $rt$\end{sideways}&
\begin{sideways}Inst\end{sideways} &
\begin{sideways}Best $\UB$\end{sideways} &
\begin{sideways}Optimal\end{sideways} &
\begin{sideways}Best $rt$\end{sideways}&
\begin{sideways}Inst\end{sideways} &
\begin{sideways}Best $\UB$\end{sideways} &
\begin{sideways}Optimal\end{sideways} &
\begin{sideways}Best $rt$\end{sideways}\\ \hline \hline
\endfirsthead
\hline
\multicolumn{12}{|l|}%
{\textbf{\tablename\ \thetable{} -- continued from previous page}}\\
\hline
999 & 999 & 999 & 555.55& 999 & 999 & 999 & 555.55& 999 & 999 & 999 & 555.55
\kill
\begin{sideways}Inst\end{sideways} &
\begin{sideways}Best $\UB$\end{sideways} &
\begin{sideways}Optimal\end{sideways} &
\begin{sideways}Best $rt$\end{sideways}&
\begin{sideways}Inst\end{sideways} &
\begin{sideways}Best $\UB$\end{sideways} &
\begin{sideways}Optimal\end{sideways} &
\begin{sideways}Best $rt$\end{sideways}&
\begin{sideways}Inst\end{sideways} &
\begin{sideways}Best $\UB$\end{sideways} &
\begin{sideways}Optimal\end{sideways} &
\begin{sideways}Best $rt$\end{sideways}\\ \hline \hline
\endhead
\hline
\multicolumn{12}{|l|}%
{Continued on next page}\\
\hline
\endfoot
\hline \hline
\endlastfoot
34 & 95 & 95 & 0.58 & 35 & 103 & 103 & 1.11 & 38 & 106 & 106 & 0.43\\
48 & 50 & 50 & 0.01 & 58 & 63 & 63 & 0.01 & 65 & 92 & 92 & 5.45\\
70 & 117 & 117 & 1.00 & 71 & 58 & \itshape 56 & 0.05 & 72 & 50 & \itshape 49 &
0.08\\
73 & 59 & \itshape 58 & 0.07 & 75 & 24 & \itshape 23 & 0.05 & 77 & 46 & 46 &
0.02\\
78 & 38 & 38 & 0.04 & 80 & 28 & \itshape 27 & 0.06 & 81 & 43 & 43 & 0.01\\
88 & 36 & 36 & 0.02 & 90 & 40 & 40 & 0.01 & 128 & 100 & 100 & 0.44\\
130 & 98 & 98 & 0.34 & 149 & 64 & 64 & 0.00 & 150 & 47 & \itshape 46 & 0.01\\
154 & 119 & 119 & 15.90 & 167 & 52 & \itshape 50 & 0.04 & 170 & 63 & 63 & 0.01\\
220 & 113 & 113 & 0.48 & 246 & 119 & 119 & 1.25 & & & &\\
\end{longtable}
\begin{longtable}{rrrr|rrrr|rrrr}
\caption{All closed instances from class \BCj{30}}
\label{tab:closed:j30}\\
999 & 999 & 999 & 555.55& 999 & 999 & 999 & 555.55& 999 & 999 & 999 & 555.55
\kill
\begin{sideways}Inst\end{sideways} &
\begin{sideways}Best $\UB$\end{sideways} &
\begin{sideways}Optimal\end{sideways} &
\begin{sideways}Best $rt$\end{sideways}&
\begin{sideways}Inst\end{sideways} &
\begin{sideways}Best $\UB$\end{sideways} &
\begin{sideways}Optimal\end{sideways} &
\begin{sideways}Best $rt$\end{sideways}&
\begin{sideways}Inst\end{sideways} &
\begin{sideways}Best $\UB$\end{sideways} &
\begin{sideways}Optimal\end{sideways} &
\begin{sideways}Best $rt$\end{sideways}\\ \hline \hline
\endfirsthead
\hline
\multicolumn{12}{|l|}%
{\textbf{\tablename\ \thetable{} -- continued from previous page}}\\
\hline
999 & 999 & 999 & 555.55& 999 & 999 & 999 & 555.55& 999 & 999 & 999 & 555.55
\kill
\begin{sideways}Inst\end{sideways} &
\begin{sideways}Best $\UB$\end{sideways} &
\begin{sideways}Optimal\end{sideways} &
\begin{sideways}Best $rt$\end{sideways}&
\begin{sideways}Inst\end{sideways} &
\begin{sideways}Best $\UB$\end{sideways} &
\begin{sideways}Optimal\end{sideways} &
\begin{sideways}Best $rt$\end{sideways}&
\begin{sideways}Inst\end{sideways} &
\begin{sideways}Best $\UB$\end{sideways} &
\begin{sideways}Optimal\end{sideways} &
\begin{sideways}Best $rt$\end{sideways}\\ \hline \hline
\endhead
\hline
\multicolumn{12}{|l|}%
{Continued on next page}\\
\hline
\endfoot
\hline \hline
\endlastfoot
4 & 104 & \itshape 101 & 1.12 & 12 & 48 & \itshape 46 & 0.04 & 13 & 63 & 63 &
0.04\\
17 & 57 & 57 & 0.02 & 20 & 32 & \itshape 31 & 0.02 & 24 & 39 & 39 & 0.02\\
32 & 114 & \itshape 113 & 0.23 & 33 & 135 & \itshape 114 & 5.70 & 37 & 119 &
\itshape 118 & 8.15\\
38 & 93 & \itshape 90 & 5.42 & 40 & 120 & \itshape 113 & 2.64 & 41 & 47 &
\itshape 46 & 0.07\\
42 & 64 & 64 & 0.02 & 45 & 56 & \itshape 54 & 0.04 & 46 & 51 & \itshape 47 &
0.06\\
47 & 46 & 46 & 0.02 & 53 & 46 & 46 & 0.02 & 57 & 70 & 70 & 0.02\\
59 & 58 & \itshape 55 & 0.09 & 60 & 47 & \itshape 46 & 0.09 & 67 & 130 & 130 &
15.89\\
68 & 174 & 174 & 27.08 & 71 & 56 & \itshape 54 & 0.08 & 75 & 65 & \itshape 61 &
0.19\\
76 & 72 & \itshape 68 & 0.19 & 77 & 48 & \itshape 46 & 0.68 & 78 & 64 & \itshape
61 & 0.07\\
79 & 71 & 71 & 0.09 & 80 & 65 & 65 & 0.03 & 89 & 80 & 80 & 0.04\\
102 & 60 & 60 & 0.04 & 114 & 42 & 42 & 0.01 & 119 & 79 & 79 & 0.02\\
123 & 151 & \itshape 150 & 265.21 & 124 & 133 & 133 & 1.72 & 129 & 145 & 145 &
0.29\\
131 & 83 & 83 & 0.02 & 133 & 101 & 101 & 0.02 & 134 & 59 & \itshape 57 & 0.14\\
138 & 96 & 96 & 0.03 & 139 & 89 & \itshape 88 & 0.03 & 144 & 102 & 102 & 0.01\\
149 & 105 & 105 & 0.01 & 154 & 134 & 134 & 34.96 & 163 & 54 & \itshape 53 & 0.15
\\
165 & 70 & \itshape 69 & 0.18 & 167 & 112 & 112 & 0.03 & 168 & 45 & \itshape 43
& 4.90\\
170 & 96 & \itshape 95 & 0.08 & 173 & 85 & 85 & 0.02 & 174 & 60 & 60 & 0.03\\
175 & 71 & \itshape 70 & 0.03 & 176 & 93 & 93 & 0.05 & 195 & 58 & \itshape 55 &
0.03\\
204 & 52 & \itshape 51 & 0.03 & 224 & 116 & 116 & 0.03 & 230 & 116 & 116 & 0.02
\\
244 & 153 & 153 & 1.24 & 247 & 175 & 175 & 0.34 & & & &\\
\end{longtable}
\begin{longtable}{rrrr|rrrr|rrrr}
\caption{All closed instances from class \BCubo{10}}
\label{tab:closed:ubo10}\\
999 & 999 & 999 & 555.55& 999 & 999 & 999 & 555.55& 999 & 999 & 999 & 555.55
\kill
\begin{sideways}Inst\end{sideways} &
\begin{sideways}Best $\UB$\end{sideways} &
\begin{sideways}Optimal\end{sideways} &
\begin{sideways}Best $rt$\end{sideways}&
\begin{sideways}Inst\end{sideways} &
\begin{sideways}Best $\UB$\end{sideways} &
\begin{sideways}Optimal\end{sideways} &
\begin{sideways}Best $rt$\end{sideways}&
\begin{sideways}Inst\end{sideways} &
\begin{sideways}Best $\UB$\end{sideways} &
\begin{sideways}Optimal\end{sideways} &
\begin{sideways}Best $rt$\end{sideways}\\ \hline \hline
\endfirsthead
\hline
\multicolumn{12}{|l|}%
{\textbf{\tablename\ \thetable{} -- continued from previous page}}\\
\hline
999 & 999 & 999 & 555.55& 999 & 999 & 999 & 555.55& 999 & 999 & 999 & 555.55
\kill
\begin{sideways}Inst\end{sideways} &
\begin{sideways}Best $\UB$\end{sideways} &
\begin{sideways}Optimal\end{sideways} &
\begin{sideways}Best $rt$\end{sideways}&
\begin{sideways}Inst\end{sideways} &
\begin{sideways}Best $\UB$\end{sideways} &
\begin{sideways}Optimal\end{sideways} &
\begin{sideways}Best $rt$\end{sideways}&
\begin{sideways}Inst\end{sideways} &
\begin{sideways}Best $\UB$\end{sideways} &
\begin{sideways}Optimal\end{sideways} &
\begin{sideways}Best $rt$\end{sideways}\\ \hline \hline
\endhead
\hline
\multicolumn{12}{|l|}%
{Continued on next page}\\
\hline
\endfoot
\hline \hline
\endlastfoot
9 & 37 & 37 & 0.00 & 16 & 28 & 28 & 0.00 & 18 & 45 & 45 & 0.00\\
23 & 32 & 32 & 0.00 & 26 & 34 & 34 & 0.00 & 29 & 33 & 33 & 0.00\\
34 & 50 & 50 & 0.01 & 38 & 57 & 57 & 0.00 & 41 & 39 & 39 & 0.00\\
43 & 40 & 40 & 0.00 & 47 & 27 & 27 & 0.00 & 58 & 31 & 31 & 0.00\\
60 & 30 & 30 & 0.00 & 75 & 32 & 32 & 0.00 & 81 & 59 & 59 & 0.00\\
\end{longtable}
\begin{longtable}{rrrr|rrrr|rrrr}
\caption{All closed instances from class \BCubo{20}}
\label{tab:closed:ubo20}\\
999 & 999 & 999 & 555.55& 999 & 999 & 999 & 555.55& 999 & 999 & 999 & 555.55
\kill
\begin{sideways}Inst\end{sideways} &
\begin{sideways}Best $\UB$\end{sideways} &
\begin{sideways}Optimal\end{sideways} &
\begin{sideways}Best $rt$\end{sideways}&
\begin{sideways}Inst\end{sideways} &
\begin{sideways}Best $\UB$\end{sideways} &
\begin{sideways}Optimal\end{sideways} &
\begin{sideways}Best $rt$\end{sideways}&
\begin{sideways}Inst\end{sideways} &
\begin{sideways}Best $\UB$\end{sideways} &
\begin{sideways}Optimal\end{sideways} &
\begin{sideways}Best $rt$\end{sideways}\\ \hline \hline
\endfirsthead
\hline
\multicolumn{12}{|l|}%
{\textbf{\tablename\ \thetable{} -- continued from previous page}}\\
\hline
999 & 999 & 999 & 555.55& 999 & 999 & 999 & 555.55& 999 & 999 & 999 & 555.55
\kill
\begin{sideways}Inst\end{sideways} &
\begin{sideways}Best $\UB$\end{sideways} &
\begin{sideways}Optimal\end{sideways} &
\begin{sideways}Best $rt$\end{sideways}&
\begin{sideways}Inst\end{sideways} &
\begin{sideways}Best $\UB$\end{sideways} &
\begin{sideways}Optimal\end{sideways} &
\begin{sideways}Best $rt$\end{sideways}&
\begin{sideways}Inst\end{sideways} &
\begin{sideways}Best $\UB$\end{sideways} &
\begin{sideways}Optimal\end{sideways} &
\begin{sideways}Best $rt$\end{sideways}\\ \hline \hline
\endhead
\hline
\multicolumn{12}{|l|}%
{Continued on next page}\\
\hline
\endfoot
\hline \hline
\endlastfoot
4 & 98 & 98 & 0.06 & 6 & 108 & 108 & 0.08 & 8 & 93 & 93 & 0.04\\
10 & 106 & 106 & 0.08 & 13 & 92 & 92 & 0.01 & 15 & 46 & \itshape 45 & 0.02\\
17 & 69 & 69 & 0.02 & 20 & 66 & \itshape 65 & 0.02 & 21 & 44 & 44 & 0.01\\
25 & 39 & 39 & 0.01 & 26 & 61 & 61 & 0.01 & 28 & 58 & 58 & 0.01\\
32 & 86 & 86 & 0.05 & 34 & 125 & 125 & 0.02 & 40 & 106 & 106 & 0.09\\
41 & 62 & 62 & 0.01 & 44 & 78 & 78 & 0.01 & 46 & 73 & 73 & 0.01\\
48 & 88 & 88 & 0.02 & 49 & 63 & 63 & 0.01 & 54 & 57 & 57 & 0.01\\
56 & 56 & 56 & 0.01 & 57 & 107 & 107 & 0.01 & 60 & 40 & 40 & 0.00\\
65 & 119 & 119 & 0.03 & 74 & 99 & 99 & 0.01 & 82 & 53 & 53 & 0.00\\
87 & 75 & 75 & 0.00 & & & & & & & &\\
\end{longtable}
\begin{longtable}{rrrr|rrrr|rrrr}
\caption{All closed instances from class \BCubo{50}}
\label{tab:closed:ubo50}\\
999 & 999 & 999 & 555.55& 999 & 999 & 999 & 555.55& 999 & 999 & 999 & 555.55
\kill
\begin{sideways}Inst\end{sideways} &
\begin{sideways}Best $\UB$\end{sideways} &
\begin{sideways}Optimal\end{sideways} &
\begin{sideways}Best $rt$\end{sideways}&
\begin{sideways}Inst\end{sideways} &
\begin{sideways}Best $\UB$\end{sideways} &
\begin{sideways}Optimal\end{sideways} &
\begin{sideways}Best $rt$\end{sideways}&
\begin{sideways}Inst\end{sideways} &
\begin{sideways}Best $\UB$\end{sideways} &
\begin{sideways}Optimal\end{sideways} &
\begin{sideways}Best $rt$\end{sideways}\\ \hline \hline
\endfirsthead
\hline
\multicolumn{12}{|l|}%
{\textbf{\tablename\ \thetable{} -- continued from previous page}}\\
\hline
999 & 999 & 999 & 555.55& 999 & 999 & 999 & 555.55& 999 & 999 & 999 & 555.55
\kill
\begin{sideways}Inst\end{sideways} &
\begin{sideways}Best $\UB$\end{sideways} &
\begin{sideways}Optimal\end{sideways} &
\begin{sideways}Best $rt$\end{sideways}&
\begin{sideways}Inst\end{sideways} &
\begin{sideways}Best $\UB$\end{sideways} &
\begin{sideways}Optimal\end{sideways} &
\begin{sideways}Best $rt$\end{sideways}&
\begin{sideways}Inst\end{sideways} &
\begin{sideways}Best $\UB$\end{sideways} &
\begin{sideways}Optimal\end{sideways} &
\begin{sideways}Best $rt$\end{sideways}\\ \hline \hline
\endhead
\hline
\multicolumn{12}{|l|}%
{Continued on next page}\\
\hline
\endfoot
\hline \hline
\endlastfoot
6 & 232 & \itshape 213 & 0.48 & 9 & 230 & \itshape 194 & 13.81 & 11 & 146 &
\itshape 141 & 0.16\\
12 & 115 & 115 & 0.13 & 13 & 134 & 134 & 0.11 & 14 & 168 & \itshape 153 & 0.10\\
15 & 105 & \itshape 99 & 0.19 & 17 & 112 & \itshape 109 & 0.19 & 18 & 164 &
\itshape 163 & 0.06\\
19 & 166 & \itshape 156 & 0.16 & 23 & 161 & 161 & 0.05 & 30 & 289 & 289 & 0.04\\
31 & 308 & \itshape 302 & 0.26 & 34 & 232 & \itshape 223 & 10.53 & 36 & 218 &
\itshape 204 & 16.38\\
37 & 232 & \itshape 229 & 0.27 & 40 & 203 & \itshape 201 & 149.02 & 41 & 139 &
139 & 0.15\\
42 & 148 & \itshape 147 & 0.06 & 43 & 100 & \itshape 98 & 0.14 & 45 & 187 &
\itshape 181 & 0.08\\
47 & 196 & 196 & 0.08 & 49 & 153 & \itshape 145 & 0.11 & 51 & 124 & 124 & 0.05\\
52 & 139 & \itshape 137 & 0.05 & 54 & 91 & \itshape 89 & 0.06 & 55 & 191 & 191 &
0.06\\
57 & 133 & \itshape 132 & 0.06 & 58 & 182 & 182 & 0.04 & 60 & 128 & 128 & 0.08\\
61 & 288 & 288 & 0.22 & 63 & 240 & 240 & 0.72 & 64 & 326 & 326 & 0.24\\
65 & 215 & \itshape 198 & 10.59 & 67 & 246 & \itshape 243 & 0.36 & 68 & 278 &
\itshape 275 & 0.41\\
71 & 156 & 156 & 0.10 & 76 & 162 & 162 & 0.09 & 77 & 260 & 260 & 0.12\\
78 & 219 & 219 & 0.10 & 80 & 298 & 298 & 0.10 & 82 & 149 & 149 & 0.04\\
84 & 169 & 169 & 0.07 & 85 & 190 & 190 & 0.04 & 87 & 269 & 269 & 0.11\\
88 & 245 & 245 & 0.05 & 89 & 218 & 218 & 0.05 & 90 & 243 & 243 & 0.06\\
\end{longtable}
\begin{longtable}{rrrr|rrrr|rrrr}
\caption{All closed instances from class \BCubo{100}}
\label{tab:closed:ubo100}\\
999 & 999 & 999 & 555.55& 999 & 999 & 999 & 555.55& 999 & 999 & 999 & 555.55
\kill
\begin{sideways}Inst\end{sideways} &
\begin{sideways}Best $\UB$\end{sideways} &
\begin{sideways}Optimal\end{sideways} &
\begin{sideways}Best $rt$\end{sideways}&
\begin{sideways}Inst\end{sideways} &
\begin{sideways}Best $\UB$\end{sideways} &
\begin{sideways}Optimal\end{sideways} &
\begin{sideways}Best $rt$\end{sideways}&
\begin{sideways}Inst\end{sideways} &
\begin{sideways}Best $\UB$\end{sideways} &
\begin{sideways}Optimal\end{sideways} &
\begin{sideways}Best $rt$\end{sideways}\\ \hline \hline
\endfirsthead
\hline
\multicolumn{12}{|l|}%
{\textbf{\tablename\ \thetable{} -- continued from previous page}}\\
\hline
999 & 999 & 999 & 555.55& 999 & 999 & 999 & 555.55& 999 & 999 & 999 & 555.55
\kill
\begin{sideways}Inst\end{sideways} &
\begin{sideways}Best $\UB$\end{sideways} &
\begin{sideways}Optimal\end{sideways} &
\begin{sideways}Best $rt$\end{sideways}&
\begin{sideways}Inst\end{sideways} &
\begin{sideways}Best $\UB$\end{sideways} &
\begin{sideways}Optimal\end{sideways} &
\begin{sideways}Best $rt$\end{sideways}&
\begin{sideways}Inst\end{sideways} &
\begin{sideways}Best $\UB$\end{sideways} &
\begin{sideways}Optimal\end{sideways} &
\begin{sideways}Best $rt$\end{sideways}\\ \hline \hline
\endhead
\hline
\multicolumn{12}{|l|}%
{Continued on next page}\\
\hline
\endfoot
\hline \hline
\endlastfoot
11 & 263 & \itshape 243 & 0.53 & 12 & 224 & \itshape 216 & 1.09 & 13 & 180 &
\itshape 158 & 0.90\\
14 & 206 & \itshape 190 & 1.90 & 16 & 144 & \itshape 131 & 16.91 & 18 & 306 &
306 & 0.26\\
19 & 200 & \itshape 177 & 4.67 & 20 & 209 & \itshape 201 & 0.36 & 21 & 262 & 262
& 0.21\\
22 & 492 & 492 & 0.30 & 27 & 225 & \itshape 204 & 0.88 & 36 & 457 & \itshape 405
& 127.34\\
38 & 483 & \itshape 477 & 3.00 & 39 & 462 & \itshape 457 & 2.60 & 41 & 363 & 363
& 0.45\\
42 & 359 & \itshape 358 & 0.44 & 44 & 491 & \itshape 483 & 0.57 & 45 & 407 & 407
& 0.38\\
46 & 283 & \itshape 278 & 1.19 & 47 & 302 & \itshape 301 & 0.32 & 48 & 433 & 433
& 0.46\\
50 & 269 & \itshape 260 & 0.81 & 51 & 272 & \itshape 267 & 0.36 & 53 & 177 & 177
& 0.22\\
55 & 247 & 247 & 0.25 & 56 & 288 & 288 & 0.27 & 57 & 356 & 356 & 0.34\\
59 & 256 & 256 & 0.21 & 60 & 188 & 188 & 0.25 & 61 & 680 & 680 & 3.59\\
62 & 540 & \itshape 526 & 17.73 & 64 & 538 & \itshape 533 & 2.30 & 65 & 451 &
\itshape 433 & 284.60\\
67 & 459 & \itshape 402 & 39.81 & 68 & 540 & \itshape 538 & 1.18 & 71 & 514 &
514 & 0.49\\
73 & 414 & \itshape 398 & 0.57 & 74 & 255 & \itshape 228 & 1.46 & 76 & 411 & 411
& 0.62\\
77 & 351 & \itshape 345 & 0.68 & 78 & 412 & \itshape 410 & 1.15 & 79 & 483 & 483
& 0.61\\
81 & 453 & \itshape 452 & 0.44 & 82 & 571 & \itshape 568 & 0.29 & 85 & 497 & 497
& 0.24\\
86 & 531 & 531 & 0.31 & 87 & 368 & \itshape 363 & 0.28 & 88 & 402 & 402 & 0.35\\
89 & 374 & 374 & 0.23 & 90 & 476 & 476 & 0.24 & & & &\\
\end{longtable}
\begin{longtable}{rrrr|rrrr|rrrr}
\caption{All closed instances from class \BCubo{200}}
\label{tab:closed:ubo200}\\
999 & 999 & 999 & 555.55& 999 & 999 & 999 & 555.55& 999 & 999 & 999 & 555.55\kill
\begin{sideways}Inst\end{sideways} &
\begin{sideways}Best UB\end{sideways} &
\begin{sideways}Optimal\end{sideways} &
\begin{sideways}Best $rt$\end{sideways}&
\begin{sideways}Inst\end{sideways} &
\begin{sideways}Best UB\end{sideways} &
\begin{sideways}Optimal\end{sideways} &
\begin{sideways}Best $rt$\end{sideways}&
\begin{sideways}Inst\end{sideways} &
\begin{sideways}Best UB\end{sideways} &
\begin{sideways}Optimal\end{sideways} &
\begin{sideways}Best $rt$\end{sideways}\\ \hline \hline
\endfirsthead
\hline
\multicolumn{12}{|l|}%
{\textbf{\tablename\ \thetable{} -- continued from previous page}}\\
\hline
999 & 999 & 999 & 555.55& 999 & 999 & 999 & 555.55& 999 & 999 & 999 & 555.55\kill
\begin{sideways}Inst\end{sideways} &
\begin{sideways}Best UB\end{sideways} &
\begin{sideways}Optimal\end{sideways} &
\begin{sideways}Best $rt$\end{sideways}&
\begin{sideways}Inst\end{sideways} &
\begin{sideways}Best UB\end{sideways} &
\begin{sideways}Optimal\end{sideways} &
\begin{sideways}Best $rt$\end{sideways}&
\begin{sideways}Inst\end{sideways} &
\begin{sideways}Best UB\end{sideways} &
\begin{sideways}Optimal\end{sideways} &
\begin{sideways}Best $rt$\end{sideways}\\ \hline \hline
\endhead
\hline
\multicolumn{12}{|l|}%
{Continued on next page}\\
\hline
\endfoot
\hline \hline
\endlastfoot
11 & 424 & \itshape 362 & 25.07 & 14 & 467 & \itshape 442 & 4.68 & 15 & 363 & \itshape 361 & 1.33\\
16 & 604 & 604 & 1.75 & 17 & 470 & 470 & 6.51 & 18 & 382 & \itshape 377 & 1.17\\
28 & 371 & 371 & 1.75 & 30 & 350 & 350 & 1.12 & 41 & 571 & \itshape 533 & 13.52\\
42 & 721 & \itshape 712 & 3.82 & 43 & 653 & \itshape 642 & 1.16 & 45 & 522 & \itshape 514 & 1.60\\
46 & 572 & 572 & 2.01 & 47 & 380 & \itshape 345 & 5.18 & 48 & 853 & 853 & 2.19\\
49 & 696 & \itshape 683 & 2.21 & 50 & 650 & 650 & 1.10 & 51 & 581 & 581 & 1.53\\
52 & 612 & 612 & 2.34 & 53 & 624 & 624 & 1.63 & 58 & 689 & 689 & 1.50\\
63 & 1424 & \itshape 1422 & 27.69 & 68 & 1205 & \itshape 1155 & 42.62 & 69 & 994 & \itshape 943 & 72.23\\
71 & 728 & \itshape 725 & 3.01 & 72 & 720 & \itshape 717 & 2.96 & 73 & 861 & \itshape 856 & 2.76\\
74 & 1176 & \itshape 1175 & 9.77 & 75 & 830 & \itshape 827 & 4.86 & 76 & 810 & \itshape 808 & 2.32\\
77 & 804 & \itshape 762 & 11.94 & 78 & 778 & \itshape 773 & 2.31 & 79 & 760 & \itshape 757 & 2.25\\
82 & 774 & 774 & 1.51 & 83 & 820 & \itshape 817 & 2.11 & 84 & 463 & 463 & 1.57\\
85 & 592 & 592 & 1.27 &  &  &  &  & &  &  & \\
\end{longtable}
\end{center}

\section{Instances with new better upper bound}

In the Table~\ref{tab:newub:all} all 51 instances are listed for which our
methods could find a new upper bound on the makespan, but could not prove the
optimality or find a optimal solution.
For each instance following parameters are given the class of the instance 
(Class), the instance number (Inst), the previously best known upper bound (Best $\UB$) on the makespan (regarding to the reported results in~\cite{rcpspmax:lib}), the new best upper bound (New $\UB$), and the lowest runtime (Best $rt$) of our methods to find a schedule with the new best upper bound.

\begin{center}
\begin{longtable}{ccccc|ccccc}
\caption{All new $\UB$ for instances from all classes}\\
\label{tab:newub:all}
\begin{sideways}Class\end{sideways} &
\begin{sideways}Inst\end{sideways} &
\begin{sideways}Best $\UB$\end{sideways} &
\begin{sideways}New $\UB$\end{sideways} &
\begin{sideways}Best $rt$\end{sideways}&
\begin{sideways}Class\end{sideways} &
\begin{sideways}Inst\end{sideways} &
\begin{sideways}Best $\UB$\end{sideways} &
\begin{sideways}New $\UB$\end{sideways} &
\begin{sideways}Best $rt$\end{sideways}\\ \hline \hline
\endfirsthead
\hline
\multicolumn{10}{|l|}%
{\textbf{\tablename\ \thetable{} -- continued from previous page}}\\
\hline
\begin{sideways}Class\end{sideways} &
\begin{sideways}Inst\end{sideways} &
\begin{sideways}Best $\UB$\end{sideways} &
\begin{sideways}New $\UB$\end{sideways} &
\begin{sideways}Best $rt$\end{sideways}&
\begin{sideways}Class\end{sideways} &
\begin{sideways}Inst\end{sideways} &
\begin{sideways}Best $\UB$\end{sideways} &
\begin{sideways}New $\UB$\end{sideways} &
\begin{sideways}Best $rt$\end{sideways}\\ \hline \hline
\endhead
\hline
\multicolumn{10}{|l|}%
{Continued on next page}\\
\hline
\endfoot
\hline \hline
\endlastfoot
\BCc & 61 & 407 & 393 & 417.240 & \BCc & 63 & 380 & 366 & 530.730\\
\BCc & 66 & 385 & 368 & 550.930 & \BCc & 67 & 367 & 350 & 151.220\\
\BCc & 69 & 388 & 380 & 80.910 & \BCc & 70 & 391 & 379 & 305.150\\
\BCc & 125 & 351 & 316 & 159.530 & \BCc & 152 & 384 & 369 & 569.710\\
\BCc & 160 & 394 & 374 & 504.980 & \BCc & 218 & 325 & 309 & 148.550\\
\BCc & 241 & 430 & 416 & 406.440 & \BCc & 245 & 468 & 453 & 483.840\\
\BCc & 310 & 277 & 267 & 549.740 & \BCc & 331 & 388 & 381 & 212.150\\
\BCc & 336 & 393 & 371 & 429.530 & \BCc & 337 & 296 & 287 & 103.620\\
\BCc & 339 & 429 & 412 & 234.700 & \BCd & 35 & 420 & 408 & 286.630\\
\BCd & 63 & 544 & 524 & 528.090 & \BCd & 158 & 696 & 687 & 390.240\\
\BCd & 246 & 489 & 463 & 406.390 & \BCj{30} & 65 & 163 & 162 & 0.150\\
\BCj{30} & 73 & 57 & 53 & 187.120 & \BCj{30} & 151 & 158 & 157 & 303.550\\
\BCubo{100} & 4 & 429 & 410 & 132.720 & \BCubo{100} & 7 & 447 & 419 & 364.420\\
\BCubo{100} & 8 & 435 & 400 & 547.800 & \BCubo{100} & 10 & 522 & 453 & 343.380\\
\BCubo{100} & 32 & 485 & 448 & 95.060 & \BCubo{100} & 33 & 435 & 418 & 313.430\\
\BCubo{100} & 34 & 488 & 425 & 245.220 & \BCubo{100} & 37 & 453 & 426 & 399.410\\
\BCubo{100} & 40 & 504 & 473 & 97.510 & \BCubo{100} & 70 & 422 & 410 & 296.430\\
\BCubo{200} & 2 & 1000 & 938 & 1029.500 & \BCubo{200} & 3 & 951 & 906 & 242.970\\
\BCubo{200} & 4 & 1009 & 963 & 1477.460 & \BCubo{200} & 5 & 866 & 852 & 278.910\\
\BCubo{200} & 6 & 939 & 841 & 311.270 & \BCubo{200} & 8 & 998 & 911 & 400.510\\
\BCubo{200} & 33 & 892 & 855 & 351.700 & \BCubo{200} & 34 & 931 & 816 & 272.330\\
\BCubo{200} & 36 & 1025 & 921 & 391.010 & \BCubo{200} & 37 & 843 & 798 & 495.620\\
\BCubo{200} & 39 & 906 & 898 & 505.510 & \BCubo{200} & 62 & 853 & 847 & 389.050\\
\BCubo{200} & 67 & 977 & 904 & 566.380 & \BCubo{200} & 70 & 1009 & 972 & 3102.610\\
\BCubo{50} & 3 & 204 & 196 & 130.220 & \BCubo{50} & 4 & 253 & 216 & 424.010\\
\BCubo{50} & 10 & 204 & 192 & 117.310 & & & & & \\
\end{longtable}\end{center}

\end{WithAppendix}
\end{document}